\documentclass[10pt,twocolumn,letterpaper]{article}

\usepackage[pagenumbers]{cvpr}           

\usepackage{graphicx}
\usepackage{amsmath}
\usepackage{amssymb}
\usepackage{booktabs}

\usepackage[pagebackref,breaklinks,colorlinks]{hyperref}

\usepackage[capitalize]{cleveref}
\crefname{section}{Sec.}{Secs.}
\Crefname{section}{Section}{Sections}
\Crefname{table}{Table}{Tables}
\crefname{table}{Tab.}{Tabs.}

\usepackage{multirow}
\usepackage{tikz}
\usetikzlibrary{shapes,arrows}
\usetikzlibrary{calc}
\usepackage{colortbl}
\usepackage[normalem]{ulem} 
\usepackage{selectp}

\usepackage{bm}

\newcommand{\modelnamePI}{$\Pi$-Nets}

\newcommand{\modelnamedense}{$\mathcal{D}$-PolyNets}  
\newcommand{\newmodelnamePI}{$\mathcal{R}$-PolyNets}

\newcommand{\cifarten}{Cifar-10}
\newcommand{\cifarhundred}{Cifar-100}
\newcommand{\stl}{STL-10}
\newcommand{\tinyimagenet}{Tiny ImageNet}
\newcommand{\imagenet}{ImageNet Large Scale Visual Recognition Challenge 2012 (ILSVRC2012)}

\providecommand{\shortpolynamesingle}{PN}
\providecommand{\shortpolyname}{\shortpolynamesingle s}
\providecommand{\shortneuralnets}{DNNs}
\providecommand{\numberDatasets}{six}

\definecolor{baseline_color}{gray}{0.5}

 \providecommand{\invar}{z}  
\providecommand{\binvar}{\bm{\invar}}  
\providecommand{\outvar}{y}  
\providecommand{\boutvar}{\bm{\outvar}}  

\definecolor{Gray}{gray}{0.95}

\usepackage[textsize=tiny]{todonotes}

\providecommand{\citep}{\cite} 
\providecommand{\citet}{\cite}

\newcommand*{\affaddr}[1]{#1} 
\newcommand*{\affmark}[1][*]
{\textsuperscript{#1}}

\def\cvprPaperID{10643} 
\def\confName{CVPR}
\def\confYear{2023}

\begin{document}

\title{Regularization of polynomial networks for image recognition}

\author{
Grigorios G Chrysos \affmark[1] \qquad Bohan Wang \affmark[1] \qquad Jiankang Deng \affmark[2] \qquad Volkan Cevher\affmark[1]\\
\affaddr{\affmark[1]LIONS, EPFL, Lausanne, Switzerland} \qquad
\affaddr{\affmark[2]Huawei UKRD} \\
{\tt\small [name.surname]@epfl.ch,}
{\tt\small jiankangdeng@gmail.com}}

\maketitle

\begin{abstract}

Deep Neural Networks (DNNs) have obtained impressive performance across tasks, however they still remain as black boxes, e.g., hard to theoretically analyze.
At the same time, Polynomial Networks (PNs) have emerged as an alternative method with a promising performance and improved interpretability but have yet to reach the performance of the powerful DNN baselines. In this work, we aim to close this performance gap. We introduce a class of \shortpolyname{}, which are able to reach the performance of ResNet across a range of  \numberDatasets{} benchmarks. We demonstrate that strong regularization is critical and conduct an extensive study of the exact regularization schemes required to match performance. To further motivate the regularization schemes, we introduce \modelnamedense{} that achieve a higher-degree of expansion than previously proposed polynomial networks. \modelnamedense{} are more parameter-efficient while achieving a similar performance as other polynomial networks. We expect that our new models can lead to an understanding of the role of elementwise activation functions (which are no longer required for training \shortpolyname). The source code is available at \url{https://github.com/grigorisg9gr/regularized_polynomials}.

\end{abstract}

\section{Introduction}
\label{sec:normalization_poly_introduction}

Deep neural networks (\shortneuralnets) are dominating the research agenda in computer vision since the previous decade owing to their stellar performance in image recognition~\citep{krizhevsky2012imagenet, he2016deep} and object detection~\cite{lin2017focal, lin2017feature}. 
The design of tailored normalization schemes~\cite{ioffe2015batch}, data augmentation~\cite{devries2017improved} and specific architectural blocks~\cite{huang2017densely, szegedy2016rethinking} have further fostered this trend. However, our theoretical understanding of \shortneuralnets{} pales in comparison. There is little progress in making \shortneuralnets{} interpretable, or a principled understanding of the training dynamics or the role of the network depth. 

\begin{figure}[tb]
    \centering
    \includegraphics[width=0.5\textwidth]{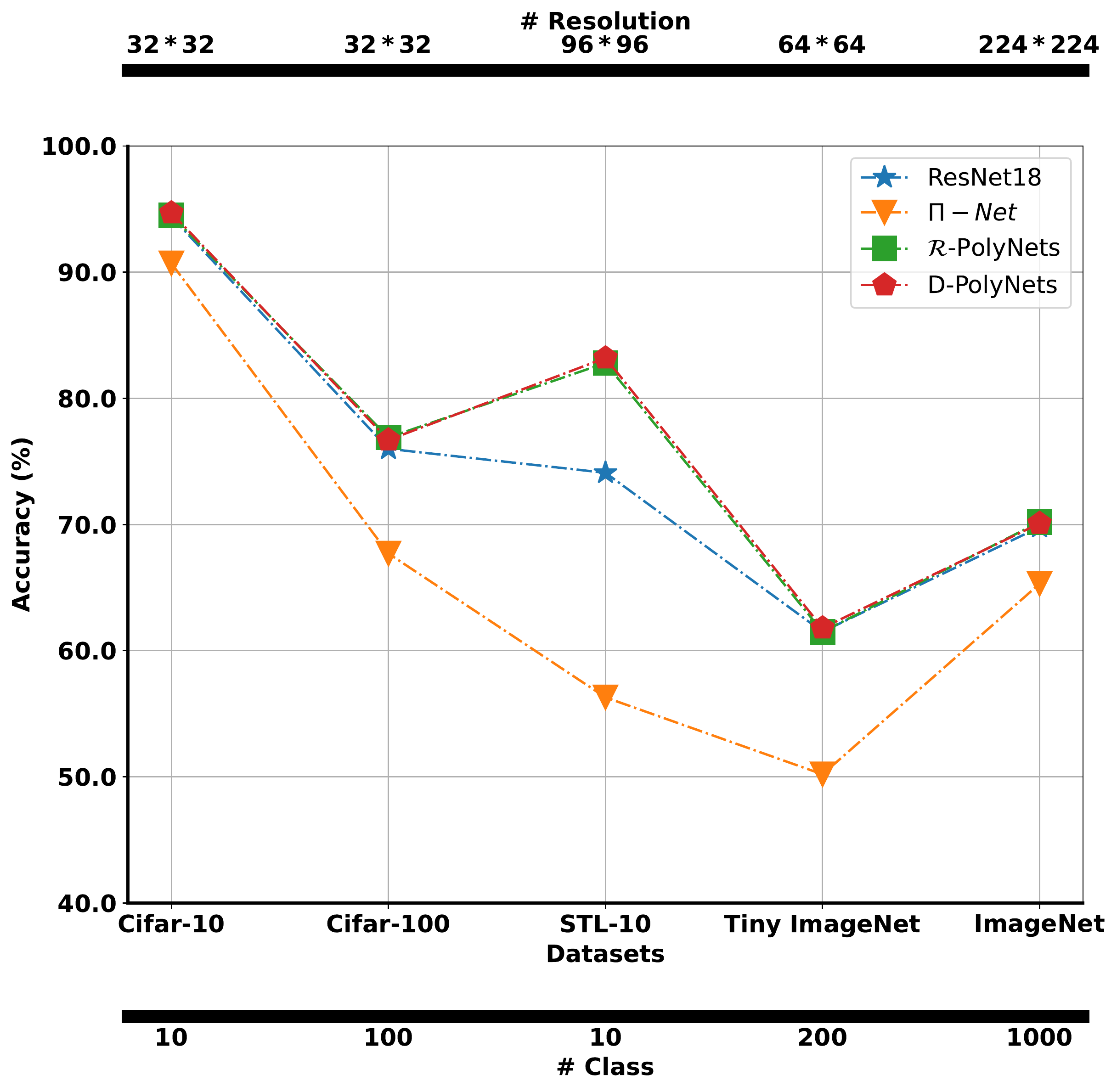}
    \caption{The proposed networks (\newmodelnamePI, \modelnamedense) enable polynomial networks to reach the performance of the powerful neural networks across a range of tasks.
    }
    \label{fig:accuracies_datasets}
\end{figure}

So far, a handful of works have attempted to mitigate that lack of understanding by designing principled architectures. 
Combining neural networks with the research on kernel methods has emerged for designing principled architectures with guarantees. 
In \cite{mairal2014convolutional}, the kernel feature map of the training data is used for achieving invariance to certain transformations. Recently, high-performing kernels were used for defining a principled architecture~\cite{simon2022reverse}. Using fixed components such as wavelets has been considered for replacing the learnable convolutions~\cite{oyallon2013generic}. 
Another approach approximates the target function with a polynomial expansion. Polynomial Nets (\shortpolyname) rely on capturing higher-order correlations of the input data for expressing the output without the use of elementwise activation functions~\cite{sejnowski1986higher}. 
Despite the progress in the principled design of networks, the aforementioned works have yet to achieve a performance comparable to standard baselines, such as the performance of the seminal residual neural networks (ResNet)~\cite{he2016deep}.

In this work, we aim to close the gap between well-established neural network architectures and principled architectures by focusing on the \shortpolyname. In particular, we concentrate on the recent parametrization of \modelnamePI~\cite{chrysos2020pinets} that has outperformed the aforementioned principled methods. We validate our hypothesis that the performance of \shortpolyname{} can be significantly improved through strong regularization schemes. To this end, we introduce a class of polynomial networks, called \newmodelnamePI. In our study, we explore which regularization schemes can improve the performance of \shortpolyname. For instance, we find that initializations proposed for neural networks~\citep{glorot2010understanding, saxe2013exact} are not optimal for \shortpolyname.
Overall, our exploration enables \newmodelnamePI{} to achieve performance on par with the (unregularized) ResNet, which is the de facto neural network baseline. 

To further motivate our regularization schemes, we design a new class of polynomial expansions achieving a higher total degree of expansion than previous \shortpolyname. In \newmodelnamePI, the final degree of expansion is obtained by a sequential concatenation of a series of lower-degree polynomial expansions. That is, \newmodelnamePI{} concatenate $N$ polynomials of second-degree to obtain a $2^N$ polynomial expansion. Instead, we use outputs from previous polynomials in the current expansion, increasing the previous total degree. Our goals are twofold: a) transfer representations from earlier polynomials, b) increase the total degree of polynomial expansion. The proposed regularization schemes are critical for training these dense polynomials, named \modelnamedense. We showcase that \modelnamedense{} are more expressive than previously proposed polynomial expansions. Overall, our contributions can be summarized as follows:\looseness-1  
\begin{itemize}
  \item We introduce a class of regularized polynomial networks, called \newmodelnamePI, in sec.~\ref{sec:normalization_poly_method}. 
  \item We propose densely connected polynomials, called  \modelnamedense. \modelnamedense{} use multiple terms from a previous polynomial as input to the current polynomial resulting in a higher-degree of expansion than previous \shortpolyname{} (sec.~\ref{sec:normalization_dense_poly_method}). 
  \item Our thorough validation in both image and audio recognition illustrates the critical components for achieving performance equivalent to vanilla \shortneuralnets. 
\end{itemize}

 \section{Related work}
\label{sec:normalization_poly_related}

\subsection{Polynomial networks}

Polynomial networks (\shortpolyname) capture higher-order interactions between the input elements using high-degree polynomial expansions. 
\shortpolyname{} have demonstrated a promising performance in standard benchmarks in image generation~\cite{chrysos2020pinets} and image recognition~\cite{chrysos2022polynomial}. 
Beyond the empirical progress, various (theoretical) properties of \shortpolyname{} have been recently explored~\cite{zhu2022controlling, pan2022study}. 
In particular,  \shortpolyname{} augment the expressivity of \shortneuralnets~\cite{fan2021expressivity}, while they offer benefits in the extrapolation or learning high-frequency functions~\cite{choraria2022the, wu2022extrapolation}. More importantly, the recent work of \cite{dubey2022scalable} highlights how \shortpolyname{} can learn powerful interpretable models. 

When \shortpolyname{} are combined with element-wise activation functions, they can achieve state-of-the-art performance as demonstrated in \cite{wang2018non, hu2018squeeze, li2019selective, chrysos2020pinets, babiloni2021poly, yin2020disentangled, yang2023polynomial}. However, many of the beneficial properties, such as the interpretability are not applicable for these models. Therefore, the hybrid models combining polynomial expansions with activation functions are not the focus of our work, since they share similar drawbacks to \shortneuralnets.

\begin{table*}[t]
\caption{Nomenclature on the symbols used in this work}
\label{tbl:notation}
\vskip 0.15in \vspace{-8mm}
\begin{center}
\begin{tabular}{|c|c|c|}
\hline Symbol & Dimension(s) & Definition \\
\hline$n, N$ & $\mathbb{N}$ & Polynomial term degree, total approximation degree. \\
$r$ & $\mathbb{N}$ & Rank of the decomposition. \\
$\bm{z}$ & $\mathbb{R}^{d}$ & Input to polynomial expansion. \\
$\bm{B}, \theta$ & $\mathbb{R}^{o \times r}, \mathbb{R}^{o}$ & Parameters in the decomposition. \\
$\boldsymbol{H}_{[n]}, \boldsymbol{J}_{[n]}, \boldsymbol{K}_{[n]}$ & $\mathbb{R}^{d \times r}, \mathbb{R}^{k \times r}, \mathbb{R}^{\omega \times r}$ & Parameters in the hierarchical decomposition. \\
$\bm{\Phi}, \bm{\Psi}$ & $\mathbb{R}^{r \times r}, \mathbb{R}^{r \times r}$ & Regularization matrices.\\
$*$ & $-$ & Hadamard (element-wise) product. \\
\hline
\end{tabular}
\end{center}
\vskip -0.1in
\end{table*}

\subsection{Regularization of neural networks}
\label{ssec:normalization_poly_related_regularization}

Deep neural networks (DNNs) can be prone to over-fitting and regularization methods are widely used to mitigate this issue. 
Hence, we summarize below three categories of regularization techniques: a) data augmentation, b) intermediate learned features and c) auxiliary loss terms.

\textbf{Data augmentation}: Data augmentation techniques are often used in image recognition pipelines~\citep{zhang2017mixup, devries2017improved, yun2019cutmix}. Mixup~\citep{zhang2017mixup} uses a linear interpolation between two training samples to improve generalization based on empirical vicinal risk minimization~\citep{chapelle2001vicinal}. Cutout~\citep{devries2017improved} removes contiguous regions from the input images, generating augmented training dataset with partially occluded versions of existing samples. This enables the network to focus on non-dominant part of the training sample. CutMix~\citep{yun2019cutmix} combines the previous two augmentation methods by replacing a patch of the image with a patch from another training image.

\textbf{Feature normalization}: Apart from the data augmentation, feature normalization enables deeper neural networks to be trained. Dropout~\citep{srivastava2014dropout} is a prominent example of such feature normalization techniques. Dropout randomly drops units (and the corresponding connections) during training to avoid co-adaptation of units. Despite its initial success, dropout has not been widely used with convolutional neural nets. Instead, dropblock~\citep{ghiasi2018dropblock} randomly drops  a contiguous region of a feature map. Additional regularization schemes, such as average or max pooling, rely on the idea of aggregating features from a local region of the feature map to avoid the sensitivity to small spatial distortions~\citep{ciregan2012multi}. 

\textbf{Auxiliary loss terms}: In addition to the classification losses, additional loss terms, such as Tikhonov regularization, can result in more stable learning and can avoid overfitting~\citep{shalev2014understanding} [Chapter 13]. Weight decay forces sparsity on the weights by penalizing their norm. \citet{cogswell2016reducing} proposes to decorrelate the different units by encouraging the covariance matrix of the features to be close to the identity matrix. For classification problems, label smoothing can prevent the networks from predicting the training examples over-confidently~\citep{muller2019does}.

Despite the progress in regularization schemes and the various theoretical connections, no consensus has been reached over a single regularization scheme that performs well in all cases, so often a combination of regularization schemes are used in modern image recognition pipelines.

 \section{Regularizing polynomial networks}
\label{sec:normalization_poly_method}

In this section, we introduce \newmodelnamePI{} in sec.~\ref{ssec:normalization_poly_proposed_model}, while we refer to the training techniques used in sec.~\ref{ssec:normalization_poly_training_techniques}.

\textbf{Notation}: Vectors (or matrices) are indicated with lowercase boldface letters e.g., $\boldsymbol{x}$ $($or $\boldsymbol{X})$. 
Tensors are identified by calligraphic letters, e.g., $\mathcal{X}$. 
The main symbols along with their dimensions are summarized in Table~\ref{tbl:notation}.

\subsection{Proposed model}
\label{ssec:normalization_poly_proposed_model}

\begin{figure}[]
\vskip 0.2in
\begin{center}
\includegraphics[width=1.0\linewidth]{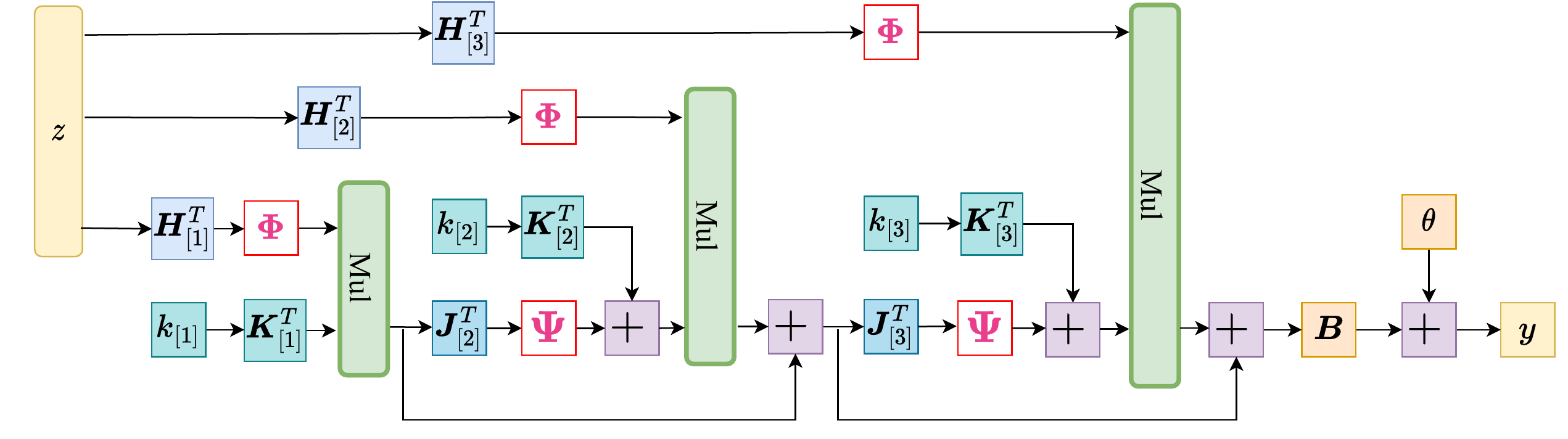}
\end{center}\vspace{-3mm}
\caption{Schematic illustration of the \newmodelnamePI{} for third-degree expansion with respect to the input $\bm{z}$ (sec.~\ref{ssec:normalization_poly_proposed_model}). The `Mul' abbreviates the Hadamard product. 
}
\label{fig:normalization_poly_schematic_network}
\vskip -0.2in
\end{figure}

One critical component for learning \shortpolyname{} is their regularization~\citep{wang2017sort}. 
To this end, we introduce a class of \shortpolyname, called \newmodelnamePI.  \newmodelnamePI{} include two regularization matrices $\bm{\Phi}, \bm{\Psi}$ that can result in different normalization schemes as we indicate below. We express an $N^{\text{th}}$ degree polynomial expansion of the input vector $\binvar$ with a simple recursive equation as follows:
\begin{equation}
\boutvar_{n}=\left(\bm{\Phi} \bm{H}_{[n]}^{T} \binvar\right) *\left(\bm{\Psi}\bm{J}_{[n]}^{T} \boutvar_{n-1}+\bm{K_{[n]}^{T}} \bm{k_{[n]}}\right) + \boutvar_{n-1},
\label{eq:normalization_poly_recursive_eq}
\end{equation}
where 
$\Big\{\bm{H_{[n]}}, \bm{J}_{[n]}, \bm{K}_{[n]}, \bm{k}_{[n]} \Big\}_{n=2}^N$ are trainable parameters. Eq. \eqref{eq:normalization_poly_recursive_eq} can be recursively applied for $n=2, \ldots, N$ for an $N^{\text{th}}$ degree polynomial expansion with $\boutvar_1 := \binvar$ and $\boutvar := \bm{B} \boutvar_N + \bm{\theta}$. The output $\boutvar$ captures high-order correlations between the input elements of $\binvar$. Eq. \eqref{eq:normalization_poly_recursive_eq} enables us to build a polynomial expansion of arbitrary degree; we demonstrate in Fig.~\ref{fig:normalization_poly_schematic_network} how this can be implemented for $3^{\text{rd}}$ degree expansion. Each term in Eq. \eqref{eq:normalization_poly_recursive_eq} has a specific purpose: a) $\bm{H}{[n]}^{T} \binvar$ performs a linear transformation of the input, b) $\bm{J}{[n]}^{T} \boutvar_{n-1}+\bm{K_{[n]}^{T}} \bm{k_{[n]}}$ performs a linear transformation of the output of the previous layer, as performed in regular neural networks. The resulting two representations are then multiplied element-wise, and a skip connection is added.\looseness-1

Our method allows for a variety of normalization schemes through the matrices $\bm{\Phi}$ and $\bm{\Psi}$. For example, we can use the matrix $\bm{\mathbb{I}} - \frac{\overrightarrow{\bm{1}}}{h}$ (where $\bm{\mathbb{I}}$ is the identity matrix, $h$ is the dimensionality of the vector being multiplied, and $\overrightarrow{\bm{1}}$ is a matrix of ones) to subtract the mean from each element, effectively creating a zero-mean vector. This extends the previously proposed \modelnamePI{} and can recover it as a special case when $\bm{\Phi}$ and $\bm{\Psi}$ are both identity transformations. However, we emphasize that normalization is necessary in our experimentation, making our method a significant extension of the previously proposed \shortpolyname{} in achieving performance on par with \shortneuralnets. We develop the details of the normalization schemes used in practice in the next section. 

We provide an additional model for our new method in case a different underlying tensor decomposition is selected, which is discussed in sec.\ref{ssec:CCP_r_pi_nets}. It is worth noting that, in practice, convolutions are often used instead of the full matrices in Eq.\ \eqref{eq:normalization_poly_recursive_eq}. This aligns with the implementation choices of our prior works\cite{chrysos2020pinets, chrysos2022polynomial}.

\subsection{Training Configuration}
\label{ssec:normalization_poly_training_techniques}
We propose a number of techniques that enforce stronger regularization of the \shortpolyname. The regularization schemes are divided into three categories: initialization (sec.~\ref{sssec:normalization_poly_init}), normalization (sec.~\ref{sssec:normalization_poly_normalization}) and auxiliary regularization (sec.~\ref{sssec:normalization_poly_regularization}). 
The precise details of each training configuration are offered in the supplementary, e.g., in Table~\ref{tbl:experiment_setting}.

\subsubsection{Initialization scheme}
\label{sssec:normalization_poly_init}

The initialization of the weights is a critical topic in deep (convolutional) neural networks~\citep{he2015delving}. The proposed initialization schemes are developed either for fully-connected neural networks~\citep{glorot2010understanding} or for convolutional neural networks~\citep{he2015delving}. However, \shortpolyname{} do not belong in the aforementioned classes of neural networks, thus we need a new initialization scheme. In our preliminary experiments, we noticed that the high-degree polynomial expansions are sensitive to the initialization scheme, and values closer to zero work better. We propose a simple initialization scheme below and defer the theoretical analysis on the initialization schemes of \shortpolyname{} to a future work.

\textbf{Technique 1} Let \begin{equation}
\bm{H}_{[n]}, \bm{J}_{[n]}, \bm{K}_{[n]} \sim \mathcal{N}\left(\mathbf{0}, \sigma^{2} \mathbf{I}\right) \text{ with } \sigma = \sqrt{\frac{D}{M_{n}}},
\end{equation}
for $n=1, \ldots, N$ with $M_{n}$ the total number of polynomial parameters of $n^{\text{th}}$ order.
In other words, we initialize polynomial parameters with zero-mean gaussian distribution. In practice, we choose $D=16$.

\subsubsection{Normalization scheme}
\label{sssec:normalization_poly_normalization}

Normalization is a core component for training of \shortneuralnets, while we expect normalization to have a significant role in enabling training \shortpolyname. Despite the popularity of batch normalization (BN)~\citep{ioffe2015batch}, BN normalizes the features across a batch of samples, which might not be ideal for high-degree expansions. Instead, instance normalization (IN)~\citep{ulyanov2016instance} computes the mean and variance for each sample and each channel to normalize the features. However, a combination of both can be beneficial for our goal. 

\textbf{Technique 2} We adopt the normalization scheme IBN, which combines instance normalization (IN) and batch normalization (BN) for \shortpolyname{} \citep{pan2018two}. 
For each block in the first three layers, we apply IN for $0.8\cdot C$ (number of channels produced by the convolution) channels, and BN for the other channels, after the first convolution. In the final layer, we only implement batch normalization to preserve discrimination between individual representations in the latent space. 
Note that the parameters $\bm{\Phi}, \bm{\Psi}$ of Eq.~\eqref{eq:normalization_poly_recursive_eq} are implemented using these normalization schemes. 

\subsubsection{Auxiliary regularization schemes}
\label{sssec:normalization_poly_regularization}
Three auxiliary regularization schemes are used: one 
auxiliary loss, one feature augmentation, and one feature regularization. Following the convention of \modelnamePI, the network consists of product of polynomials of the form of Eq.~\eqref{eq:normalization_poly_recursive_eq}. That is, the output of each polynomial is used as the input to the next polynomial, which results in a higher degree polynomial expansion, increasing the significance of regularization.  We utilize 
Label smoothing~\citep{muller2019does}, DropBlock~\citep{ghiasi2018dropblock} and max pooling. Max pooling is applied after each polynomial expansion except the final one.

\textbf{Technique 3} We adopt 
Label smoothing \citep{muller2019does}, DropBlock \citep{ghiasi2018dropblock} and max pooling in the proposed framework. 
Label smoothing is applied on the labels and Dropblock on the feature maps.  We add max pooling layer after each individual polynomial expansion (of the form of Eq.~\eqref{eq:normalization_poly_recursive_eq}) in \newmodelnamePI{}.

\section{Dense connections across polynomials}
\label{sec:normalization_dense_poly_method}

To showcase the representative power of the regularized polynomial expansion,
we propose a new type of \shortpolyname{}, called \modelnamedense{}. In the polynomial networks proposed above (or in the literature), a sequence of polynomial expansions is used with
the output of $i^{\text{th}}$ polynomial being used as the input of  $(i+1)^{\text{th}}$ polynomial. Adding $N$ such second-degree polynomials results in an overall polynomial expansion of degree $2^{{N}}$. 

\begin{figure}[]
\vskip -0.22in
\begin{center}
\includegraphics[width=0.88\linewidth]{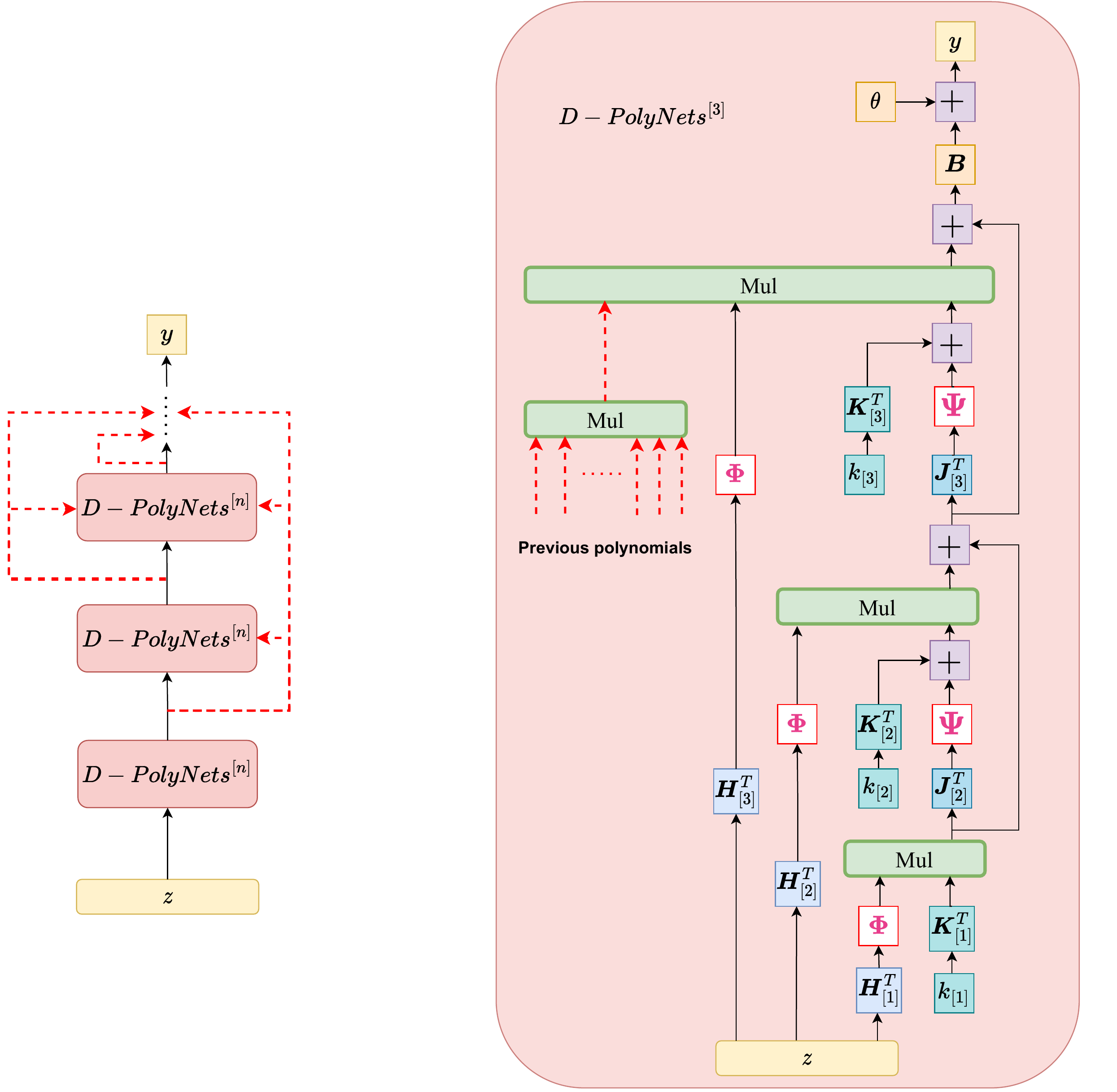}
\end{center}\vspace{-3mm}
\caption{Schematic illustration of \modelnamedense{}. On the left the overall structure is presented, while on the right a single third-degree polynomial using the structure of \modelnamedense{} is visualized. The red arrows depict the newly added connections with respect to previous polynomial expansions. 
}
\label{fig:schematic_dense_poly}
\vskip -0.2in
\end{figure}

In \modelnamedense{}, we enable additional connections across polynomials which results in a higher degree of expansion. 
To achieve that we enable outputs from the $i^{\text{th}}$ polynomial being used as a) input to the next polynomial, b) as a term in the Hadamard products of a next polynomial. 
Let us assume that each polynomial includes a single recursive step with potentially multiple terms. In Eq.~\eqref{eq:normalization_poly_recursive_eq}, taking a single recursive step (i.e., $n=2$) includes a Hadamard product between a filtered version of the input $\binvar$ and a filtered version of the previous recursive term $\boutvar_{1}$. On the contrary, in \modelnamedense, a single recursive term includes both of the aforementioned terms along with the outputs from previous polynomials. 
The schematic in Fig.~\ref{fig:schematic_dense_poly} depicts \modelnamedense{} assuming each polynomial includes a single recursive step. This can be trivially extended to any number of recursive steps, while each polynomial can also rely on a different tensor decomposition.

If we denote as $\boutvar^{[i]}$ the output of the $i^{\text{th}}$ polynomial, then the recursive formulation of \modelnamedense{} based on \eqref{eq:normalization_poly_recursive_eq} is the following expressions: 
\begin{equation}
\begin{split}
    \boutvar_{n}^{[i]}=\boutvar_{n-1}^{[i]} + \\
    \left(\bm{\Phi} \bm{H}_{[n]}^{T} \binvar\right) *\left(\bm{\Psi}\bm{J}_{[n]}^{T} \boutvar_{n-1}+\bm{K_{[n]}^{T}} \bm{k_{[n]}}\right) *_{\tau=1}^{i-1} \boutvar^{[\tau]}.
\end{split}
\label{eq:normalization_dense_poly_recursive_eq}
\end{equation}
The total degree of expansion in Eq.~\eqref{eq:normalization_dense_poly_recursive_eq} is higher than the corresponding one in \newmodelnamePI{} or \modelnamePI. This makes the requirement for strong regularization imperative to avoid  exploding gradients.

Equation \eqref{eq:normalization_poly_recursive_eq} enables us to build a regularized polynomial expansion of arbitrary degree, $\boutvar_{n}$. Apart from the training techniques in sec~\ref{ssec:normalization_poly_training_techniques}, we propose below specific techniques to enforce a strong regularization of \modelnamedense{}.

\textbf{Training Configuration of \modelnamedense{}:} Our preliminary experiments indicate that iterative normalization~\citep{huang2019iterative} is beneficial in this case. Additionally, we include a learnable parameter $\rho_{\tau}$ which regularizes the contribution of each previous polynomial in the current Hadamard product.

\section{Experiments}
\label{sec:normalization_poly_experiments}

In this section, we evaluate the proposed models across a range of \numberDatasets{} in image recognition and one standard dataset in audio recognition. We describe below the datasets, and the setup, then we conduct an ablation study to evaluate the contribution of different techniques in sec.~\ref{ssec:normalization_poly_experiment_ablation}. Sequentially, we conduct the main body of the experiments in various widely-used datasets in sec.~\ref{ssec:normalization_poly_main_experiment_image_recognition} and sec.~\ref{sec:imagenet}. We extend beyond the standard image classification tasks, with audio classification and fine-grained classification in sec.~\ref{ssec:audio_classification} and sec.~\ref{ssec:oxford_flower_classification} respectively. Details on the datasets along with additional experiments (including an experiment with deeper networks and the runtime comparison in FLOPs) are developed in sec.~\ref{sec:normalization_poly_additional_experiments_suppl}. The results in the supplementary verify the empirical findings below, while the the proposed \newmodelnamePI{} has a similar number of FLOPs as the \modelnamePI, e.g., in Table~\ref{tbl:imagenet_flops}.

\begin{figure}[tb]
    \centering
    \includegraphics[width=0.46\textwidth]{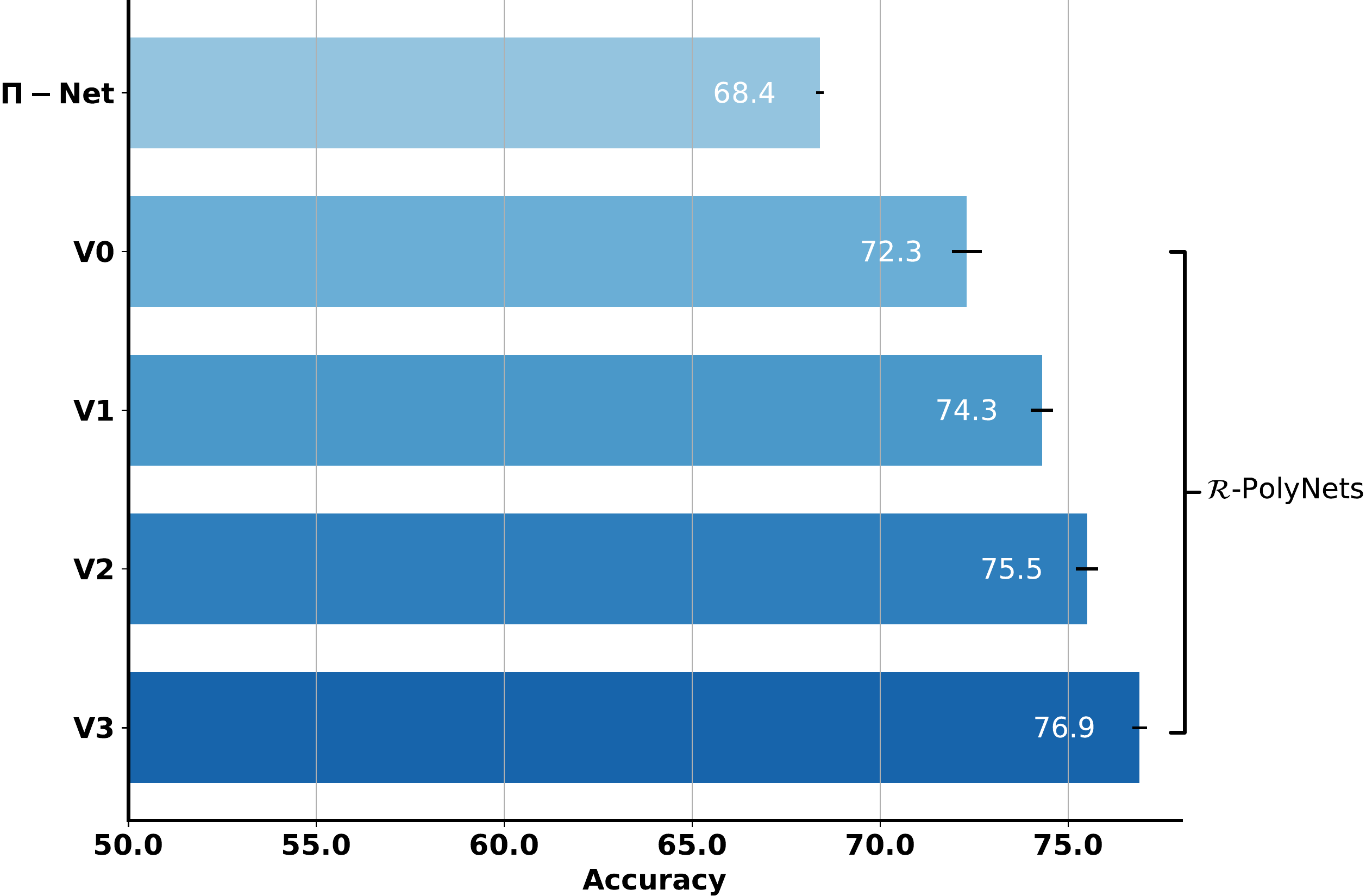} 
    \caption{
    Accuracy on \cifarhundred{}. The symbols `V0', `V1', `V2' and `V3' denote \newmodelnamePI{} (IBN), \newmodelnamePI{} (IBN + max pooling), \newmodelnamePI{} (IBN + max pooling + Dropblock) and \newmodelnamePI{} (IBN + max pooling + Dropblock + Label smoothing) respectively. Note that the normalization scheme adds a significant improvement, and similarly the regularized loss (i.e., Label smoothing) 
    has also a considerable effect. Overall, training techniques such as Dropblock and Label smoothing improve the testing performance and obtain a result comparable to the  baseline model (vanilla ResNet18).
    }
    \label{fig:design_r_pi_net}
\end{figure}

\textbf{Training details}: Each model is trained for $120$ epochs with batch size $128$. The SGD optimizer is used with initial learning rate of $0.1$. The learning rate
is multiplied with a factor of $0.1$ in epochs $40, 60, 80, 100$. For \tinyimagenet{}, the learning rates of \newmodelnamePI{} and \modelnamedense{} are multiplied by $0.92$ every epoch. For data augmentation we adopt random cropping with 4-pixel padding and horizontal flipping. Unless mentioned otherwise, each experiment is conducted \textbf{5 times} and the average and the standard deviations are also reported.\looseness-1

\textbf{Compared methods}: The family of \modelnamePI{} is the main baseline we use in our comparisons. Namely, the \modelnamePI{} use the structure of ResNet18, where each block from the original ResNet is converted into a second-degree polynomial expansion. As a reminder, \modelnamePI{} do not use element-wise activation functions, and result in a high-degree polynomial expansion. The recent PDC~\cite{chrysos2022polynomial} is also added as a baseline with many of its properties shared with \modelnamePI. We also report the accuracy of two further methods: (a) the popular ResNet18, which is the de facto baseline in image recognition, (b) hybrid \modelnamePI{}, i.e., polynomial expansions with element-wise activation functions. The last two methods are added as a reference (thus added with grayscale color). Outperforming the vanilla ResNet18 or the hybrid \modelnamePI{} is not our goal in this work. We aim at demonstrating for the first time that polynomial expansions can be on par with feed-forward neural networks.

\subsection{Ablation study}
\label{ssec:normalization_poly_experiment_ablation}

Below, we conduct three ablation experiments. In the first experiment, we showcase how the different components proposed in sec.~\ref{sec:normalization_poly_method} can decrease the error in image recognition. We choose \cifarhundred{} for this experiment. 
To facilitate the presentation, we gradually insert different regularization components on \modelnamePI{} to advocate for stronger regularization techniques and highlight their benefit in the final accuracy. 

Fig.~\ref{fig:design_r_pi_net} summarizes the results of the experiment. Notice that the initialization plus the normalization scheme already makes a significant impact on the accuracy. Then, 
max pooling, the feature augmentation and regularized loss contribute to reduce overfitting and to achieve the final performance. 
In the next experiments, we consider the last row of Fig.~\ref{fig:design_r_pi_net} as the model that is used for the main comparisons.

\begin{table}[t]
\caption{
Accuracy of \newmodelnamePI{} (IBN + max pooling + Dropblock + Label smoothing) and \modelnamedense{} (IBN + max pooling + Dropblock + Label smoothing) with different initialization schemes on \cifarhundred{}. Note that \modelnamedense{} contains $7$M parameters (down from the $11$M of \newmodelnamePI) and one block less than \newmodelnamePI. 
} 
\label{tbl:normalization_poly_ablation_results_initialization}
\vskip 0.15in \vspace{-6mm}
\begin{center}
\begin{tabular}{lll}
\multicolumn{1}{c}{\bf Model}  &\multicolumn{1}{c}{\bf Initialization} &\multicolumn{1}{c}{\bf Accuracy}
\\ \hline
\multirow{3}{*}{\newmodelnamePI{}} & \multicolumn{1}{l}{Xavier} &  $0.765 \pm 0.002$ \\ 
& Orthogonal & $0.765 \pm 0.001$ \\
& Kaiming normal & $0.767 \pm 0.003$         \\
& Kaiming uniform & $0.767 \pm 0.004$        \\
& zero-mean & $\bm{0.769 \pm 0.002}$        \\
\cline{1-3}
\multirow{3}{*}{\modelnamedense{}} & \multicolumn{1}{l}{Xavier} &  $0.761 \pm 0.004$ \\ 
& Orthogonal & $0.764 \pm 0.001$ \\
& Kaiming normal & $0.764 \pm 0.002$        \\
& Kaiming uniform & $0.760 \pm 0.004$        \\
& zero-mean & $\bm{0.767 \pm 0.003}$        \\
\cline{1-3}

\end{tabular}
\end{center}
\vskip -0.1in
\end{table}

In the second experiment, we utilize well-established 
initialization schemes, i.e., Xavier initialization~\citep{glorot2010understanding}, orthogonal matrix initialization~\citep{saxe2013exact}, Kaiming initialization~\citep{he2015delving},  and evaluate their performances on the proposed \newmodelnamePI. 
The results in Table~\ref{tbl:normalization_poly_ablation_results_initialization} indicate that previously proposed initializations cannot perform as well in the \newmodelnamePI. This is not contradictory to the studies of those initialization schemes, since they were derived for the neural network structure, which differs substantially from the structure of \shortpolyname.

\begin{figure}[tb]
    \centering\vspace{-5mm}
    \includegraphics[width=0.475\textwidth]{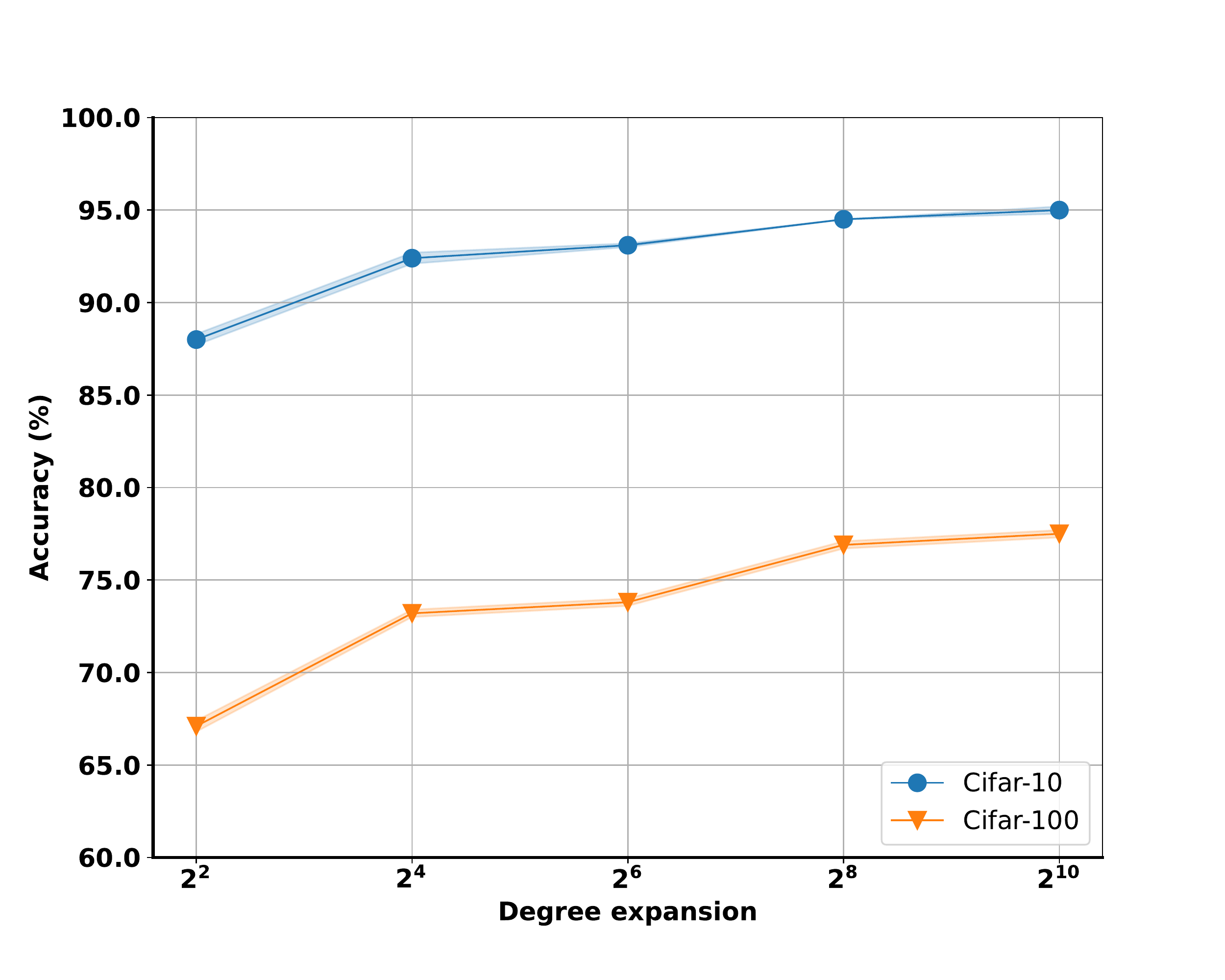}
    \caption{Accuracy of \newmodelnamePI{} with varying degree polynomials on \cifarten{} and \cifarhundred{}. 
    }
    \label{fig:varying_order_r_pi_net_plot}
\end{figure}

In the third experiment, we vary the degree of polynomials applied in \newmodelnamePI{}. The results in Fig.~\ref{fig:varying_order_r_pi_net_plot} indicate that an increasing degree of polynomial expansion can increase the accuracy. Given the correspondence between the standard ResNet18 with $8$ residual blocks and the $2^8$ degree polynomial, we use this $2^8$ degree in the rest of the experiments unless explicitly stated otherwise.

\subsection{Image classification on smaller datasets}
\label{ssec:normalization_poly_main_experiment_image_recognition}

\begin{figure*}[htbp]
    \centering
    \subfloat[\cifarten]{\includegraphics[width=0.32\linewidth]{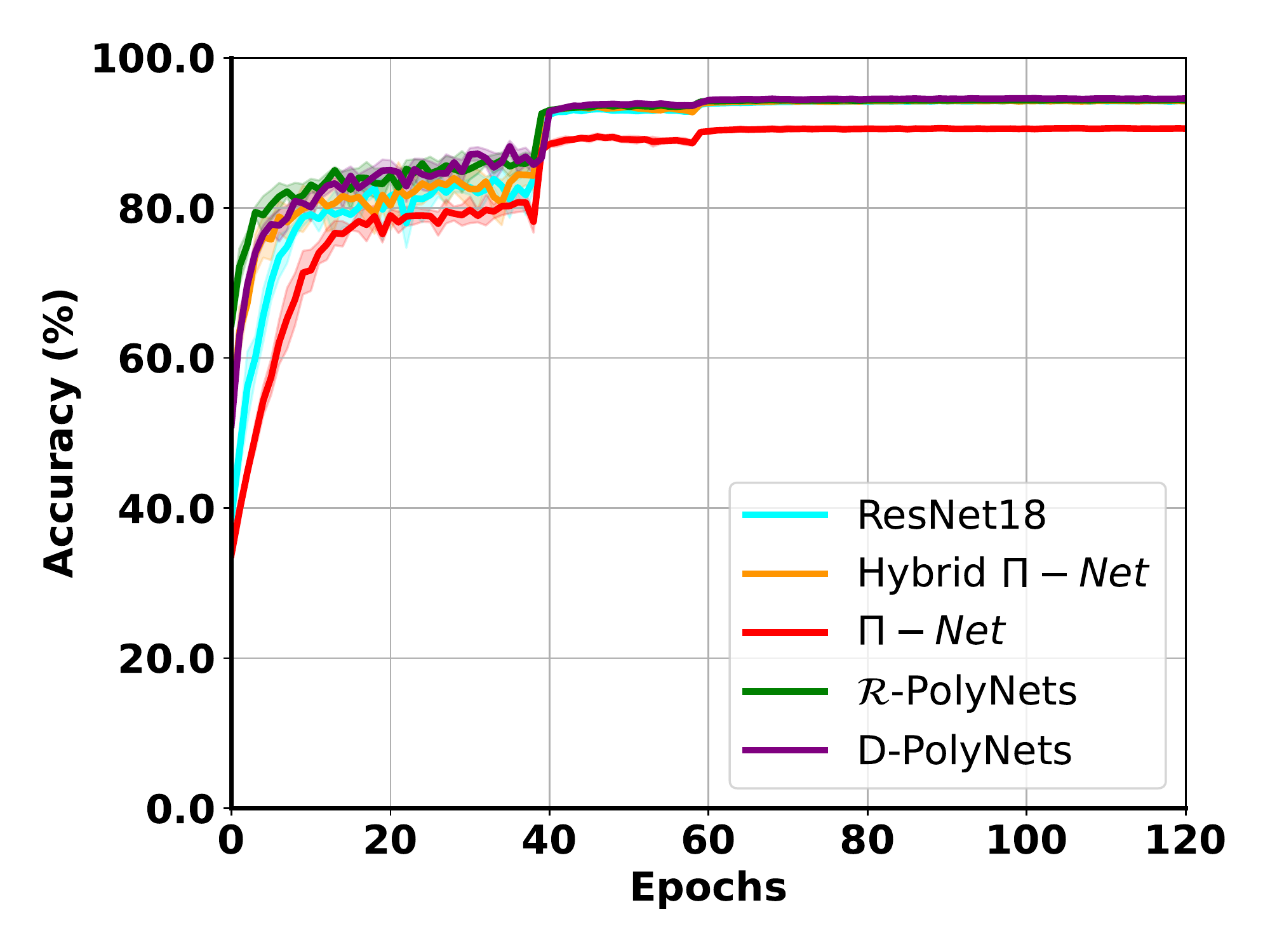}}
    \subfloat[\cifarhundred]{\includegraphics[width=0.32\textwidth]{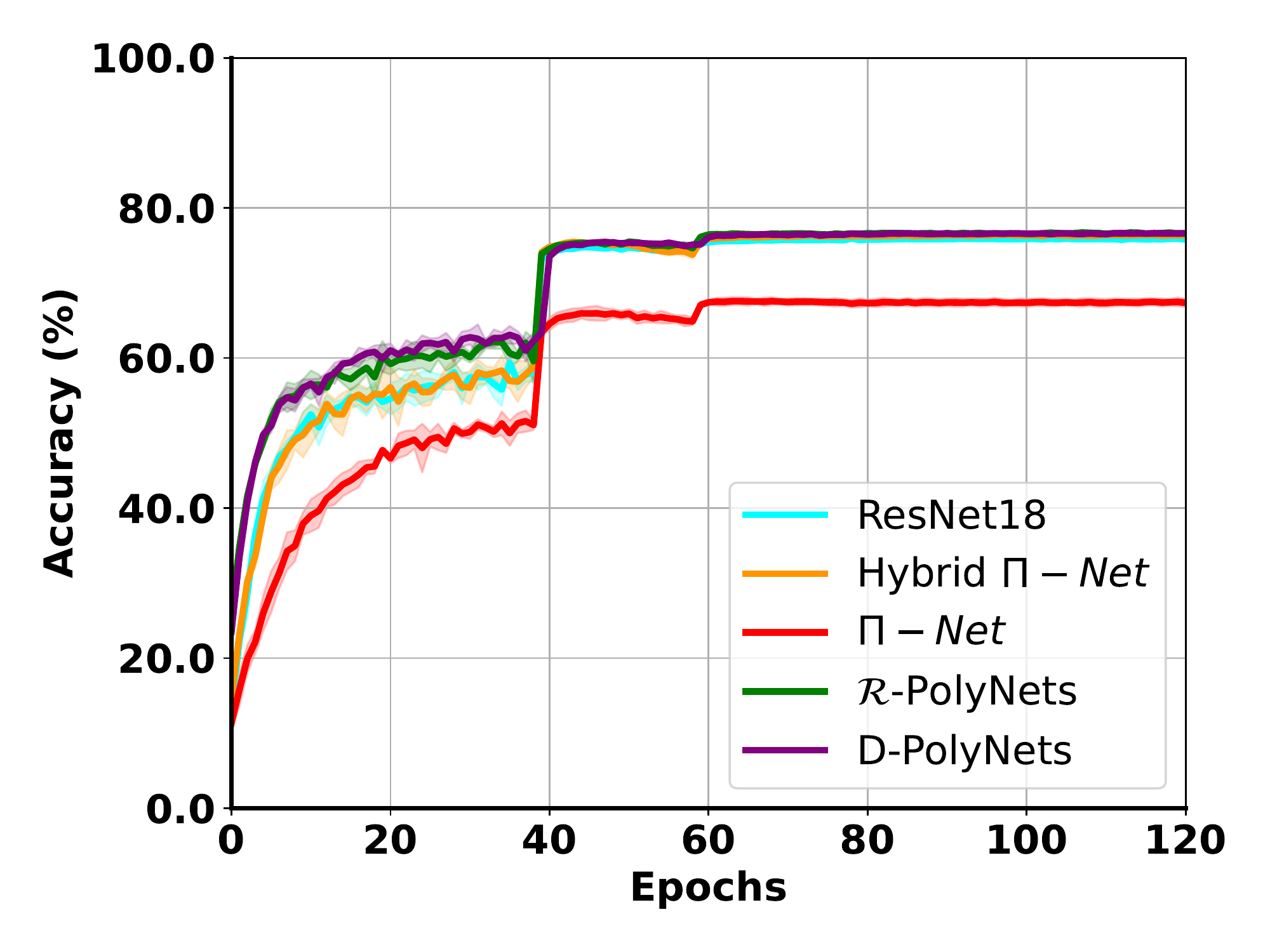}}
    \subfloat[\stl]{\includegraphics[width=0.32\textwidth]{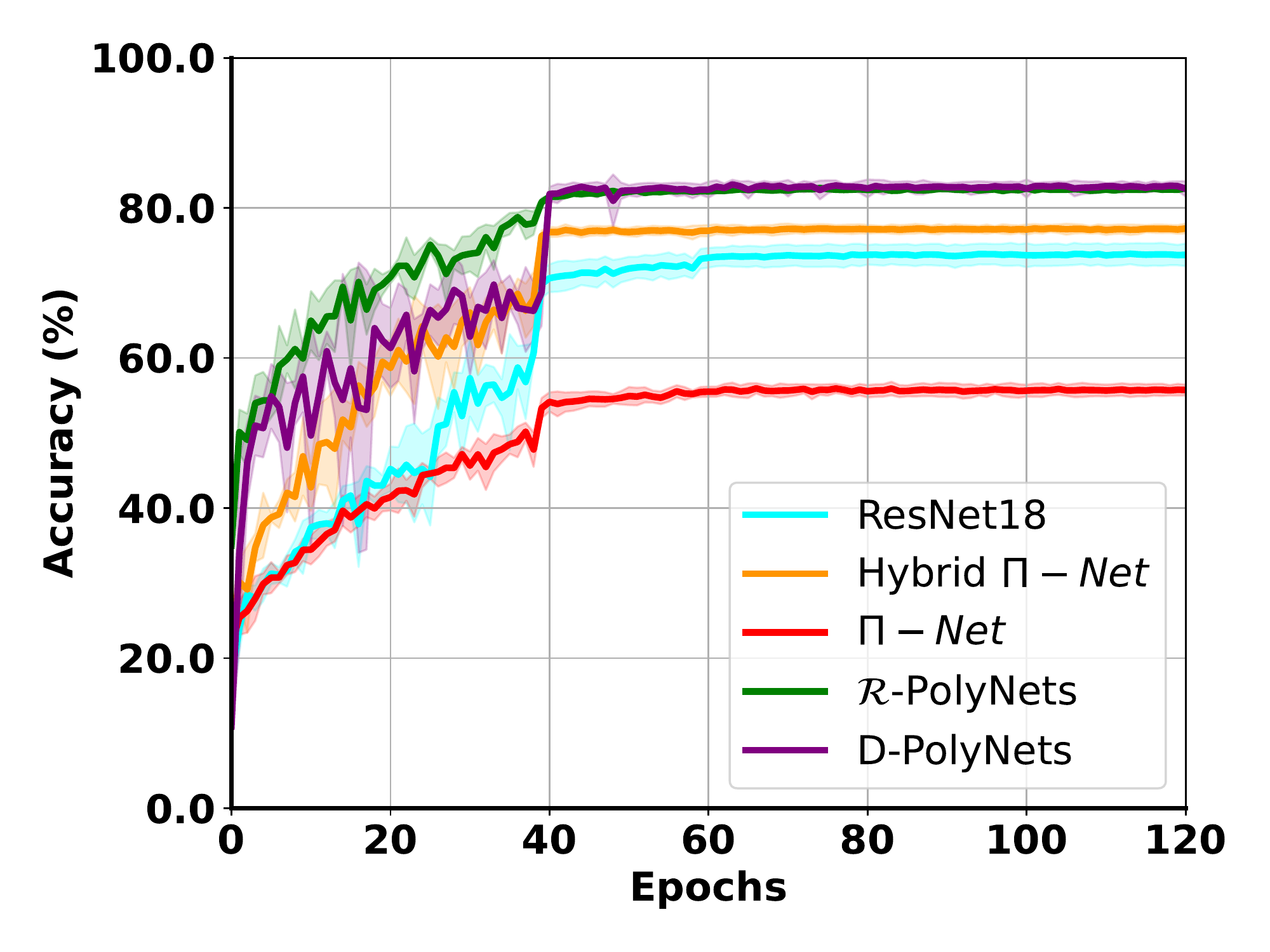}}
    \caption{
    Test error on (a) \cifarten, (b) \cifarhundred{} and (c) \stl{}. The highlighted region depicts the variance in the accuracy. Interestingly, the proposed \newmodelnamePI{} and \modelnamedense{} outperform the \modelnamePI{} from the first few epochs, while the absolute difference in the error is not decreasing as the training progresses. Notice that the proposed training techniques enable \newmodelnamePI{} and \modelnamedense{} to be on par with the ResNet18 baseline in \stl{} and even outperform the baselines in \cifarhundred.  
    }
    \label{fig:normalization_poly_error_per_epoch}
\end{figure*}

We conduct our main experimentation in the following four datasets: \cifarten, \cifarhundred, \stl, \tinyimagenet. Table~\ref{tbl:results} exhibits the accuracy of each compared method across all datasets. The results indicate that the \newmodelnamePI{} consistently outperform the baseline \modelnamePI{} by a large margin. In \stl{} the accuracy increases from $56.3\%$ to $82.8\%$, which is a \emph{$47.1\%$ relative increase} in the performance. In Fig.~\ref{fig:normalization_poly_error_per_epoch} the test accuracy per epoch is depicted; notice that the proposed method has a consistently higher accuracy over the \modelnamePI{} throughout the training. This consistent improvement demonstrates the efficacy of the proposed method. 
In Table~\ref{tbl:results}, ResNet18 and hybrid \modelnamePI{} are added. \newmodelnamePI{} achieve a higher performance than ResNet18 and hybrid \modelnamePI{} on the three benchmarks of \cifarten{}, \cifarhundred{} and \stl, while the three methods perform on par on \tinyimagenet. These observations verify our proposal that regularized polynomials can achieve a similar performance with the standard baselines of neural networks or hybrid \modelnamePI.

\begin{table}[t]
\caption{Accuracy on \cifarten{}, \cifarhundred{}, \stl{} and \tinyimagenet{}. The symbol ‘\# par’ abbreviates the number of parameters. \modelnamedense{} containing 7M parameters. Note that \newmodelnamePI{} and \modelnamedense{} without activation functions can outperform \modelnamePI{} without activation functions significantly on \cifarten{}, \cifarhundred{}, \stl{} and \tinyimagenet{}. Moreover, \newmodelnamePI{} and \modelnamedense{}  can match the performances of baseline models (e.g. \modelnamePI{} with activation functions and ResNet18) on \cifarten{}, \cifarhundred{} and \stl{}. 
} 
\label{tbl:results}
\vskip 0.15in \vspace{-6mm}
\begin{center}
\small
\begin{tabular}{llll}
\multicolumn{1}{c}{\bf Dataset}  &\multicolumn{1}{c}{\bf Model} &\multicolumn{1}{c}{\bf \# par} &\multicolumn{1}{c}{\bf Accuracy}
\\ \hline
\multirow{3}{*}{\cifarten{}} &
\multicolumn{1}{l}{\textcolor{baseline_color}{ResNet18}}   & \textcolor{baseline_color}{$11.2$M} & \multicolumn{1}{l}{\textcolor{baseline_color}{$0.944 \pm 0.001$}} \\
&   \multicolumn{1}{l}{\textcolor{baseline_color}{Hybrid \modelnamePI{}}} & \textcolor{baseline_color}{$6.0$M}  & \multicolumn{1}{l}{\textcolor{baseline_color}{$0.944 \pm 0.002$}}                 \\\cline{2-4}
& PDC & $5.4$M & $0.909 \pm 0.002$        \\
& \multicolumn{1}{l}{\modelnamePI{}} & $11.9$M  & $0.907\pm0.003$ \\ 
& \newmodelnamePI{} & $11.9$M  & $0.945 \pm 0.000$ \\
& \modelnamedense{} & $7.1$M & $\bm{0.947 \pm 0.002}$        \\
\cline{1-4}
\multirow{3}{*}{\cifarhundred{}} &
\multicolumn{1}{l}{\textcolor{baseline_color}{ResNet18}}               & \textcolor{baseline_color}{$11.2$M}            & \multicolumn{1}{l}{\textcolor{baseline_color}{$0.760 \pm 0.003$}} \\
& \multicolumn{1}{l}{\textcolor{baseline_color}{Hybrid \modelnamePI{}}}     &     \textcolor{baseline_color}{$6.1$M}                   & \multicolumn{1}{l}{\textcolor{baseline_color}{$0.765 \pm 0.004$}}                       \\\cline{2-4} 
& PDC & $5.5$M & $0.689 \pm 0.002$        \\
& \multicolumn{1}{l}{\modelnamePI{}} & $11.9$M  & $0.677\pm0.006$ \\ 
& \newmodelnamePI{} & $11.9$M  & $\bm{0.769 \pm 0.002}$ \\
& \modelnamedense{} & $7.2$M &  $0.767 \pm 0.003$        \\  

\cline{1-4}
\multirow{3}{*}{\stl{}} &
\multicolumn{1}{l}{\textcolor{baseline_color}{ResNet18}}                &   \textcolor{baseline_color}{$11.2$M}   & \multicolumn{1}{l}{\textcolor{baseline_color}{$0.741 \pm 0.016$}} \\
& \multicolumn{1}{l}{\textcolor{baseline_color}{Hybrid \modelnamePI{}}}        &   \textcolor{baseline_color}{$6.0$M}             &     \multicolumn{1}{l}{\textcolor{baseline_color}{$0.775 \pm 0.006$}}                  \\\cline{2-4} 
& PDC & $5.4$M &   ${0.681\pm0.006}$     \\
& \multicolumn{1}{l}{\modelnamePI{}} & $11.9$M  & ${0.563\pm0.008}$ \\ 
& \newmodelnamePI{} & $11.9$M  & $0.828 \pm 0.003$ \\
& \modelnamedense{} & $7.1$M &  $\bm{0.834 \pm 0.006}$        \\  

\cline{1-4}
\multirow{3}{*}{\tinyimagenet{}} &
\multicolumn{1}{l}{\textcolor{baseline_color}{ResNet18}}                  &   \textcolor{baseline_color}{$11.3$M}      & \multicolumn{1}{l}{\textcolor{baseline_color}{$0.615 \pm 0.002$}} \\ 
&  \multicolumn{1}{l}{\textcolor{baseline_color}{Hybrid \modelnamePI{}}}    & \textcolor{baseline_color}{$6.1$M} &   \multicolumn{1}{l}{\textcolor{baseline_color}{$0.611 \pm 0.004$}}                  \\ \cline{2-4} 
& PDC & $5.5$M & $0.452 \pm 0.002$       \\
& \multicolumn{1}{l}{\modelnamePI{}} & $12.0$M  & $0.502 \pm 0.007$ \\ 
& \newmodelnamePI{} & $12.0$M  & $0.615 \pm 0.004$ \\
& \modelnamedense{} & $7.2$M &  $\bm{0.618 \pm 0.001}$        \\  

\cline{1-4}

\end{tabular}
\end{center}
\vskip -0.1in
\end{table}

\begin{table*}[t]
\caption{\textbf{ImageNet classification results.} We compare our models with state-of-the-art deep convolutional neural networks, \modelnamePI{} and hybrid \modelnamePI{}. We report the top-1 and top-5 accuracy on the validation set of ImageNet as well as the number of parameters. Our models are highlighted in gray. The symbol ‘\# par’ abbreviates the number of parameters.}
\label{tbl:imagenet_results}
\vskip 0.15in \vspace{-8mm}
\begin{center}
\small
\begin{tabular}{llllll}
\hline Model & Image Size & \# par $(\mathrm{M})$ & Top-1 Acc. $(\%)$ & Top-5 Acc. $(\%)$\\
\hline ImageNet-1K trained models & & & & \\
\hline ResNet18 & $224^2$ & $11.69$ & $69.758$ & $89.078$ \\
ResNet18 without activations & $224^2$ & $11.69$ & $20.536$ & $39.986$ \\
Hybrid \modelnamePI{} & $224^2$ & $11.96$ & $\bm{70.740}$ & $\bm{89.548}$ \\
\modelnamePI{} & $224^2$ &  $12.38$  & $65.280$ & $85.958$ \\
\rowcolor{Gray}
\newmodelnamePI{} & $224^2$ & $12.38$  & $70.228$  & $89.390$\\
\rowcolor{Gray}
\modelnamedense{} & $224^2$ & $11.36$  & $70.090$ & $89.424$  \\
\hline
\end{tabular}
\end{center}
\end{table*}

\textbf{Discussion on \newmodelnamePI}: A reasonable question would be whether the studied training techniques are unique in enabling the training of \newmodelnamePI. Our preliminary experiments with alternative methods indicate that different combinations of training techniques can indeed perform well. However, the proposed techniques are the only ones we found that perform well in a range of datasets. To verify this, we conducted two experiments to assess alternative training techniques of \newmodelnamePI{}. 

In the first experiment, we increase the weight decay to train \newmodelnamePI{} on \cifarhundred{}.
We notice that max pooling and Label smoothing can prevent \newmodelnamePI{} from overfitting the train samples. However, the alternative regularization schemes may also help \newmodelnamePI{} achieve the similar final performance. To verify this, we conduct another experiment of training \newmodelnamePI{} with different regularization techniques. The result is presented in Table~\ref{tbl:different_regularization_scheme}. 

As the results  in Table~\ref{tbl:ablation_results_different_weight_decay} and Table~\ref{tbl:different_regularization_scheme} illustrate, different combinations of regularization techniques can indeed improve the test performance of \newmodelnamePI. Notice that the CutMix and Stochastic depth can also help \shortpolynamesingle{} without element-wise activation functions perform on par with established \shortneuralnets{}. However, we find their behavior is dataset-dependent, so we do not include them in our final scheme. 

\textbf{Discussion on \modelnamedense}: The results in Table~\ref{tbl:results} demonstrate that \modelnamedense{} can match the test performance of \newmodelnamePI{}, which both outperform the baseline \modelnamePI{}. However, \modelnamedense{} have $41.7\%$ fewer parameters. The representation power of \modelnamedense{} is improved, because the total degree of polynomial expansion output by \modelnamedense{} can reach $\binvar^{588}$ higher than that of \modelnamePI{} ($\binvar^{256}$) when they use at most $8$ blocks each. The results validate our assumption that \modelnamedense{} are more  expressive.\looseness-1

\begin{table}[]
\caption{The impact of weight decay changes on \newmodelnamePI{} (IBN + max pooling + Dropblock + Label smoothing) trained on \cifarhundred{}. 
}
\label{tbl:ablation_results_different_weight_decay}
\begin{center}
\vspace{-4mm}
\begin{tabular}{|l|l|}
\hline
Weight decay & Accuracy \\ \hline
  {$5e^{-4}$}      &    {$0.766 \pm 0.002$}           \\ \hline
  $6e^{-4}$      &     $0.768 \pm 0.004$        \\ \hline
  $7e^{-4}$   &        $\bm{0.768 \pm 0.002}$          \\ \hline

\end{tabular}
\end{center}\vspace{-3mm}
\end{table}

\begin{table}[t]
\caption{Accuracy on \cifarhundred{}. Note that Data aumentation (i.e. Cutmix) can achieve the best test performance. Overall, these alternative training techniques obtain a state-of-the-art result with respect to the baseline model (ResNet18). 
}
\label{tbl:different_regularization_scheme}
\vskip 0.15in
\begin{center}
\vspace{-6mm}
\scalebox{0.63}{
\begin{tabular}{|l|l|}
\hline
Models & Accuracy \\ \hline
    {\newmodelnamePI{} (IBN + maxpooling + Dropblock + Label smoothing)} & {$0.766 \pm 0.002$} \\ \hline
      \newmodelnamePI{} (IBN + maxpooling + Dropblock + CutMix) & \textbf{$\bm{0.771 \pm 0.002}$}\\ \hline
      \newmodelnamePI{} (IBN + maxpooling + Stochastic depth + Label smoothing) & $0.769 \pm 0.002$\\ \hline

\end{tabular}
}
\end{center}
\vskip -0.1in
\end{table}

\subsection{ImageNet Classification}
\label{sec:imagenet}

We conduct a large-scale classification
experiment on ImageNet~\citep{deng2009imagenet}.
We employ the mmclassification toolkit \cite{2020mmclassification} to train networks on the training set and report the $224 \times 224$ single-crop top-1 and the top-5 errors on the validation set. Our pre-processing and augmentation strategy follows the settings of the baseline (\ie, ResNet18). All models are trained for 100 epochs on 8  
GPUs with 32 images per GPU (effective batch size of 256) with synchronous SGD of momentum 0.9. The learning rate is initialized to 0.1, and decays by a factor of 10 at the $30^{\text{th}}$, $60^{\text{th}}$, and $90^{\text{th}}$ epochs. The results in Table~\ref{tbl:imagenet_results} validate our findings on the rest of the datasets and confirm that \newmodelnamePI{} are able to reach the performance of standard neural network baselines. \modelnamedense{} perform on par with \newmodelnamePI, while having a slightly reduced number of parameters. The reduction in parameters is smaller than the respective numbers with smaller scale datasets (e.g., Table~\ref{tbl:results}) , which might indicate that further regularization is required for \modelnamedense{} when scaled to more complex datasets.

\subsection{Audio classification}
\label{ssec:audio_classification}

We perform an experiment on the Speech Commands dataset to evaluate \newmodelnamePI{} and \modelnamedense{} on a distribution that differs from that of natural images. 
The accuracy for each model is reported in Table~\ref{tbl:results_speech}. Noticeably, hybrid \modelnamePI{}, \newmodelnamePI{}, \modelnamedense{} and ResNet18 can achieve the accuracy over $0.975$, which showcases the representative power of \newmodelnamePI{} and \modelnamedense{} on audio classification. By comparison, \modelnamePI{} and PDC has accuracy below $0.975$.

\begin{table}[t]
\caption{Accuracy on Speech Command. Note the Hybrid \modelnamePI{}, \newmodelnamePI{}, \modelnamedense{} and ResNet18 can achieve the same test performance.}
\label{tbl:results_speech}
\vskip 0.15in \vspace{-7mm}
\begin{center}
\begin{tabular}{llll}
\multicolumn{1}{c}{\bf Dataset}  &\multicolumn{1}{c}{\bf Model} &\multicolumn{1}{c}{\bf \# par} &\multicolumn{1}{c}{\bf Accuracy}
\\ \hline
\multirow{4}{*}{Speech command}     & \multicolumn{1}{l}{\textcolor{baseline_color}{ResNet18}}  & \textcolor{baseline_color}{$11.2$M} &  \multicolumn{1}{l}{\textcolor{baseline_color}{$0.977$}} \\
&   \multicolumn{1}{l}{\textcolor{baseline_color}{Hybrid \modelnamePI{}}} & \textcolor{baseline_color}{$6.0$M}                 &   \multicolumn{1}{l}{\textcolor{baseline_color}{$0.977$}}               \\ \cline{2-4}
& PDC & $5.4$M & $0.972$                 \\ 
& \modelnamePI{} &  $11.9$M &         $0.972$      \\
                           & \newmodelnamePI{} & $11.9$M  & $0.977$                 \\
& \modelnamedense{} & $7.2$M & $0.977$                 \\ 
 \cline{1-4} 
\end{tabular}
\end{center}
\end{table}

\subsection{Fine-grained classification}
\label{ssec:oxford_flower_classification}

We conduct one last experiment on fine-grained classification to validate further the regularization scheme of \newmodelnamePI. We select the Oxford $102$ Flowers dataset~\citep{nilsback2008automated}, which contains $102$ flower categories with $10$ training images and $10$ validation images annotated per class.  
The accuracy for each model is exhibited in Table~\ref{tbl:results_fine_grained_classification_oxford_flower}, where \newmodelnamePI{} performs favorably to the vanilla ResNet18. As a reminder, the goal of our experiments is not to demonstrate state-of-the-art behavior, but rather to focus on achieving performance at least comparable to the existing \shortneuralnets.

\begin{table}[ht]
\caption{Accuracy on Oxford 102 Flowers. Note \newmodelnamePI{} and \modelnamedense{} perform favorably to the vanilla ResNet18.}
\label{tbl:results_fine_grained_classification_oxford_flower}
\vskip 0.15in \vspace{-7mm}
\begin{center}
\begin{tabular}{llll}
\multicolumn{1}{c}{\bf Dataset}  &\multicolumn{1}{c}{\bf Model} &\multicolumn{1}{c}{\bf \# par} &\multicolumn{1}{c}{\bf Accuracy}
\\ \hline
\multirow{4}{*}{Oxford Flower}     & \multicolumn{1}{l}{\textcolor{baseline_color}{ResNet18}}  & \textcolor{baseline_color}{$11.2$M} &  \multicolumn{1}{l}{\textcolor{baseline_color}{$0.877$}} \\
&   \multicolumn{1}{l}{\textcolor{baseline_color}{Hybrid \modelnamePI{}}} & \textcolor{baseline_color}{$6.1$M}                 &   \multicolumn{1}{l}{\textcolor{baseline_color}{$0.889$}}               \\ \cline{2-4}
& PDC & $5.5$M & $0.885$                 \\ 
& \modelnamePI{} &  $11.9$M &       $0.826$        \\
                           & \newmodelnamePI{} & $11.9$M  & $\bm{0.949}$                 \\
& \modelnamedense{} & $7.2$M & $0.941$                 \\ 

 \cline{1-4} 
\end{tabular}
\end{center}
\end{table}

 \vspace{-3mm}
\section{Conclusion}
\label{sec:normalization_poly_conclusion}
In this work, we focus on Polynomial Nets (\shortpolyname) for image recognition. We propose a new parametrization of \shortpolyname{} that enables them to avoid reported overfitting issues using custom initialization and normalization schemes. We showcase how the proposed model, called \newmodelnamePI, extends previously proposed \shortpolynamesingle{} models. Our thorough evaluation with \numberDatasets{} datasets exhibits a significant improvement over previously proposed \shortpolynamesingle{} models and establish \newmodelnamePI{} as an alternative to existing \shortneuralnets. Furthermore, we introduce \modelnamedense{} that leverage dense connections across sequential polynomials to capture higher-order correlations. Experimentally, \modelnamedense{} verify their expressivity over alternative \shortpolyname. 
We believe that our work can encourage further research in alternative models for image recognition.

\noindent\textbf{Limitations}: A deeper theoretical understanding of \shortpolyname{} is needed, particularly regarding the link between the degree, the regularization requirements and the generalization error. Concretely, each block of polynomials we are using is composed of lower-degree polynomial expansions. We hypothesize that the high-degree obtained from the sequential polynomial blocks might be sufficient for image recognition tasks, but might not suffice for harder tasks. In addition, the theoretical study of the initialization or the regularization requirements on \shortpolyname{} remains elusive. 

 \section*{Acknowledgements}
\label{sec:acks}

We are thankful to the reviewers for their feedback and constructive comments. This work was supported by Hasler Foundation Program: Hasler Responsible AI (project number 21043). This project has received funding from the European Research Council (ERC) under the European Union's Horizon 2020 research and innovation programme (grant agreement n° 725594 - time-data).  
{\small
\bibliographystyle{ieee_fullname}
\bibliography{egbib}

\begin{thebibliography}{10}\itemsep=-1pt

\bibitem{babiloni2021poly}
Francesca Babiloni, Ioannis Marras, Filippos Kokkinos, Jiankang Deng,
  Grigorios~G Chrysos, and Stefanos Zafeiriou.
\newblock Poly-nl: Linear complexity non-local layers with polynomials.
\newblock In {\em International Conference on Computer Vision (ICCV)}, 2021.

\bibitem{chapelle2001vicinal}
Olivier Chapelle, Jason Weston, L{\'e}on Bottou, and Vladimir Vapnik.
\newblock Vicinal risk minimization.
\newblock In {\em Advances in neural information processing systems (NeurIPS)},
  pages 416--422, 2001.

\bibitem{choraria2022the}
Moulik Choraria, Leello~Tadesse Dadi, Grigorios~G Chrysos, Julien Mairal, and
  Volkan Cevher.
\newblock The spectral bias of polynomial neural networks.
\newblock In {\em International Conference on Learning Representations (ICLR)},
  2022.

\bibitem{chrysos2022polynomial}
Grigorios~G Chrysos, Markos Georgopoulos, Jiankang Deng, Jean Kossaifi, Yannis
  Panagakis, and Anima Anandkumar.
\newblock Augmenting deep classifiers with polynomial neural networks.
\newblock In {\em European Conference on Computer Vision (ECCV)}, 2022.

\bibitem{chrysos2020pinets}
Grigorios~G Chrysos, Stylianos Moschoglou, Giorgos Bouritsas, Yannis Panagakis,
  Jiankang Deng, and Stefanos Zafeiriou.
\newblock $\pi-$nets: Deep polynomial neural networks.
\newblock In {\em Conference on Computer Vision and Pattern Recognition
  (CVPR)}, 2020.

\bibitem{ciregan2012multi}
Dan Ciregan, Ueli Meier, and J{\"u}rgen Schmidhuber.
\newblock Multi-column deep neural networks for image classification.
\newblock In {\em Conference on Computer Vision and Pattern Recognition
  (CVPR)}, pages 3642--3649. IEEE, 2012.

\bibitem{coates2011analysis}
Adam Coates, Andrew Ng, and Honglak Lee.
\newblock An analysis of single-layer networks in unsupervised feature
  learning.
\newblock In {\em Proceedings of the fourteenth international conference on
  artificial intelligence and statistics}, pages 215--223. JMLR Workshop and
  Conference Proceedings, 2011.

\bibitem{cogswell2016reducing}
Michael Cogswell, Faruk Ahmed, Ross Girshick, Larry Zitnick, and Dhruv Batra.
\newblock Reducing overfitting in deep networks by decorrelating
  representations.
\newblock In {\em International Conference on Learning Representations (ICLR)},
  2016.

\bibitem{2020mmclassification}
MMClassification Contributors.
\newblock Openmmlab's image classification toolbox and benchmark.
\newblock \url{https://github.com/open-mmlab/mmclassification}, 2020.

\bibitem{deng2009imagenet}
Jia Deng, Wei Dong, Richard Socher, Li-Jia Li, Kai Li, and Li Fei-Fei.
\newblock Imagenet: A large-scale hierarchical image database.
\newblock In {\em Conference on Computer Vision and Pattern Recognition
  (CVPR)}, pages 248--255, 2009.

\bibitem{devries2017improved}
Terrance DeVries and Graham~W Taylor.
\newblock Improved regularization of convolutional neural networks with cutout.
\newblock {\em arXiv preprint arXiv:1708.04552}, 2017.

\bibitem{dubey2022scalable}
Abhimanyu Dubey, Filip Radenovic, and Dhruv Mahajan.
\newblock Scalable interpretability via polynomials.
\newblock In {\em Advances in neural information processing systems (NeurIPS)},
  2022.

\bibitem{fan2021expressivity}
Feng-Lei Fan, Mengzhou Li, Fei Wang, Rongjie Lai, and Ge Wang.
\newblock Expressivity and trainability of quadratic networks.
\newblock {\em arXiv preprint arXiv:2110.06081}, 2021.

\bibitem{ghiasi2018dropblock}
Golnaz Ghiasi, Tsung-Yi Lin, and Quoc~V Le.
\newblock Dropblock: A regularization method for convolutional networks.
\newblock {\em arXiv preprint arXiv:1810.12890}, 2018.

\bibitem{glorot2010understanding}
Xavier Glorot and Yoshua Bengio.
\newblock Understanding the difficulty of training deep feedforward neural
  networks.
\newblock In {\em International Conference on Artificial Intelligence and
  Statistics (AISTATS)}, pages 249--256, 2010.

\bibitem{he2015delving}
Kaiming He, Xiangyu Zhang, Shaoqing Ren, and Jian Sun.
\newblock Delving deep into rectifiers: Surpassing human-level performance on
  imagenet classification.
\newblock In {\em International Conference on Computer Vision (ICCV)}, pages
  1026--1034, 2015.

\bibitem{he2016deep}
Kaiming He, Xiangyu Zhang, Shaoqing Ren, and Jian Sun.
\newblock Deep residual learning for image recognition.
\newblock In {\em Proceedings of the IEEE conference on computer vision and
  pattern recognition}, pages 770--778, 2016.

\bibitem{hu2018squeeze}
Jie Hu, Li Shen, and Gang Sun.
\newblock Squeeze-and-excitation networks.
\newblock In {\em Proceedings of the IEEE conference on computer vision and
  pattern recognition}, pages 7132--7141, 2018.

\bibitem{huang2017densely}
Gao Huang, Zhuang Liu, Laurens Van Der~Maaten, and Kilian~Q Weinberger.
\newblock Densely connected convolutional networks.
\newblock In {\em Conference on Computer Vision and Pattern Recognition
  (CVPR)}, pages 4700--4708, 2017.

\bibitem{huang2019iterative}
Lei Huang, Yi Zhou, Fan Zhu, Li Liu, and Ling Shao.
\newblock Iterative normalization: Beyond standardization towards efficient
  whitening.
\newblock In {\em Conference on Computer Vision and Pattern Recognition
  (CVPR)}, pages 4874--4883, 2019.

\bibitem{ioffe2015batch}
Sergey Ioffe and Christian Szegedy.
\newblock Batch normalization: Accelerating deep network training by reducing
  internal covariate shift.
\newblock In {\em International conference on machine learning}, pages
  448--456. PMLR, 2015.

\bibitem{krizhevsky2009learning}
Alex Krizhevsky, Geoffrey Hinton, et~al.
\newblock Learning multiple layers of features from tiny images.
\newblock 2009.

\bibitem{cifar100}
Alex Krizhevsky, Vinod Nair, and Geoffrey Hinton.
\newblock Cifar-100 (canadian institute for advanced research).

\bibitem{krizhevsky2012imagenet}
Alex Krizhevsky, Ilya Sutskever, and Geoffrey~E Hinton.
\newblock Imagenet classification with deep convolutional neural networks.
\newblock In {\em Advances in neural information processing systems (NeurIPS)},
  pages 1097--1105, 2012.

\bibitem{le2015tiny}
Ya Le and Xuan Yang.
\newblock Tiny imagenet visual recognition challenge.
\newblock {\em CS 231N}, 7:7, 2015.

\bibitem{li2019selective}
Xiang Li, Wenhai Wang, Xiaolin Hu, and Jian Yang.
\newblock Selective kernel networks.
\newblock In {\em Conference on Computer Vision and Pattern Recognition
  (CVPR)}, pages 510--519, 2019.

\bibitem{lin2017feature}
Tsung-Yi Lin, Piotr Doll{\'a}r, Ross Girshick, Kaiming He, Bharath Hariharan,
  and Serge Belongie.
\newblock Feature pyramid networks for object detection.
\newblock In {\em Conference on Computer Vision and Pattern Recognition
  (CVPR)}, pages 2117--2125, 2017.

\bibitem{lin2017focal}
Tsung-Yi Lin, Priya Goyal, Ross Girshick, Kaiming He, and Piotr Doll{\'a}r.
\newblock Focal loss for dense object detection.
\newblock In {\em International Conference on Computer Vision (ICCV)}, pages
  2980--2988, 2017.

\bibitem{mairal2016end}
Julien Mairal.
\newblock End-to-end kernel learning with supervised convolutional kernel
  networks.
\newblock {\em Advances in neural information processing systems (NeurIPS)},
  29, 2016.

\bibitem{mairal2014convolutional}
Julien Mairal, Piotr Koniusz, Zaid Harchaoui, and Cordelia Schmid.
\newblock Convolutional kernel networks.
\newblock {\em Advances in neural information processing systems (NeurIPS)},
  27, 2014.

\bibitem{muller2019does}
Rafael M{\"u}ller, Simon Kornblith, and Geoffrey Hinton.
\newblock When does label smoothing help?
\newblock {\em arXiv preprint arXiv:1906.02629}, 2019.

\bibitem{nilsback2008automated}
Maria-Elena Nilsback and Andrew Zisserman.
\newblock Automated flower classification over a large number of classes.
\newblock In {\em 2008 Sixth Indian Conference on Computer Vision, Graphics \&
  Image Processing}, pages 722--729. IEEE, 2008.

\bibitem{oyallon2013generic}
Edouard Oyallon, St{\'e}phane Mallat, and Laurent Sifre.
\newblock Generic deep networks with wavelet scattering.
\newblock {\em arXiv preprint arXiv:1312.5940}, 2013.

\bibitem{pan2022study}
Chao Pan and Chuanyi Zhang.
\newblock On the study of sample complexity for polynomial neural networks.
\newblock {\em arXiv preprint arXiv:2207.08896}, 2022.

\bibitem{pan2018two}
Xingang Pan, Ping Luo, Jianping Shi, and Xiaoou Tang.
\newblock Two at once: Enhancing learning and generalization capacities via
  ibn-net.
\newblock In {\em Proceedings of the European Conference on Computer Vision
  (ECCV)}, pages 464--479, 2018.

\bibitem{saxe2013exact}
Andrew~M Saxe, James~L McClelland, and Surya Ganguli.
\newblock Exact solutions to the nonlinear dynamics of learning in deep linear
  neural networks.
\newblock In {\em International Conference on Learning Representations (ICLR)},
  2014.

\bibitem{sejnowski1986higher}
Terrence~J Sejnowski.
\newblock Higher-order boltzmann machines.
\newblock In {\em AIP Conference Proceedings}, volume 151, pages 398--403.
  American Institute of Physics, 1986.

\bibitem{selvaraju2017grad}
Ramprasaath~R Selvaraju, Michael Cogswell, Abhishek Das, Ramakrishna Vedantam,
  Devi Parikh, and Dhruv Batra.
\newblock Grad-cam: Visual explanations from deep networks via gradient-based
  localization.
\newblock In {\em International Conference on Computer Vision (ICCV)}, pages
  618--626, 2017.

\bibitem{shalev2014understanding}
Shai Shalev-Shwartz and Shai Ben-David.
\newblock {\em Understanding machine learning: From theory to algorithms}.
\newblock Cambridge university press, 2014.

\bibitem{simon2022reverse}
James~Benjamin Simon, Sajant Anand, and Mike Deweese.
\newblock Reverse engineering the neural tangent kernel.
\newblock In {\em International Conference on Machine Learning (ICML)}, pages
  20215--20231, 2022.

\bibitem{srivastava2014dropout}
Nitish Srivastava, Geoffrey Hinton, Alex Krizhevsky, Ilya Sutskever, and Ruslan
  Salakhutdinov.
\newblock Dropout: a simple way to prevent neural networks from overfitting.
\newblock {\em The journal of machine learning research}, 15(1):1929--1958,
  2014.

\bibitem{szegedy2016rethinking}
Christian Szegedy, Vincent Vanhoucke, Sergey Ioffe, Jon Shlens, and Zbigniew
  Wojna.
\newblock Rethinking the inception architecture for computer vision.
\newblock In {\em Conference on Computer Vision and Pattern Recognition
  (CVPR)}, pages 2818--2826, 2016.

\bibitem{ulyanov2016instance}
Dmitry Ulyanov, Andrea Vedaldi, and Victor Lempitsky.
\newblock Instance normalization: The missing ingredient for fast stylization.
\newblock {\em arXiv preprint arXiv:1607.08022}, 2016.

\bibitem{wang2018non}
Xiaolong Wang, Ross Girshick, Abhinav Gupta, and Kaiming He.
\newblock Non-local neural networks.
\newblock In {\em Conference on Computer Vision and Pattern Recognition
  (CVPR)}, pages 7794--7803, 2018.

\bibitem{wang2017sort}
Yan Wang, Lingxi Xie, Chenxi Liu, Siyuan Qiao, Ya Zhang, Wenjun Zhang, Qi Tian,
  and Alan Yuille.
\newblock Sort: Second-order response transform for visual recognition.
\newblock In {\em International Conference on Computer Vision (ICCV)}, pages
  1359--1368, 2017.

\bibitem{warden2018speech}
Pete Warden.
\newblock Speech commands: A dataset for limited-vocabulary speech recognition.
\newblock {\em arXiv preprint arXiv:1804.03209}, 2018.

\bibitem{wu2022extrapolation}
Yongtao Wu, Zhenyu Zhu, Fanghui Liu, Grigorios~G Chrysos, and Volkan Cevher.
\newblock Extrapolation and spectral bias of neural nets with hadamard product:
  a polynomial net study.
\newblock In {\em Advances in neural information processing systems (NeurIPS)},
  2022.

\bibitem{xie2017aggregated}
Saining Xie, Ross Girshick, Piotr Doll{\'a}r, Zhuowen Tu, and Kaiming He.
\newblock Aggregated residual transformations for deep neural networks.
\newblock In {\em Conference on Computer Vision and Pattern Recognition
  (CVPR)}, pages 1492--1500, 2017.

\bibitem{yang2023polynomial}
Guandao Yang, Sagie Benaim, Varun Jampani, Kyle Genova, Jonathan~T Barron,
  Thomas Funkhouser, Bharath Hariharan, and Serge Belongie.
\newblock Polynomial neural fields for subband decomposition and manipulation.
\newblock In {\em Advances in neural information processing systems (NeurIPS)},
  2022.

\bibitem{yin2020disentangled}
Minghao Yin, Zhuliang Yao, Yue Cao, Xiu Li, Zheng Zhang, Stephen Lin, and Han
  Hu.
\newblock Disentangled non-local neural networks.
\newblock In {\em European Conference on Computer Vision (ECCV)}, pages
  191--207. Springer, 2020.

\bibitem{yun2019cutmix}
Sangdoo Yun, Dongyoon Han, Seong~Joon Oh, Sanghyuk Chun, Junsuk Choe, and
  Youngjoon Yoo.
\newblock Cutmix: Regularization strategy to train strong classifiers with
  localizable features.
\newblock In {\em International Conference on Computer Vision (ICCV)}, pages
  6023--6032, 2019.

\bibitem{zhang2017mixup}
Hongyi Zhang, Moustapha Cisse, Yann~N Dauphin, and David Lopez-Paz.
\newblock mixup: Beyond empirical risk minimization.
\newblock In {\em International Conference on Learning Representations (ICLR)},
  2018.

\bibitem{zhu2022controlling}
Zhenyu Zhu, Fabian Latorre, Grigorios~G Chrysos, and Volkan Cevher.
\newblock Controlling the complexity and lipschitz constant improves polynomial
  nets.
\newblock In {\em International Conference on Learning Representations (ICLR)},
  2022.

\end{thebibliography}
}
\clearpage 

\newpage
\appendix
\section*{Contents of the Appendix}
\label{sec:appendixes}

The following sections are included in the appendix:
\begin{itemize}
    \item A review of the \modelnamePI{} is in sec.~\ref{ssec:recurive_pi_nets}. 
    \item An alternative parameterization to \newmodelnamePI{} is introduced in sec.~\ref{ssec:CCP_r_pi_nets}. 
    \item A number of auxiliary tables and visualizations that could not fit in the main paper are in sec.~\ref{sec:normalization_poly_aux_tables_suppl}.
    \item Lastly, a number of additional experiments are conducted in sec.~\ref{sec:normalization_poly_additional_experiments_suppl}. 
\end{itemize}

\section{Background: \modelnamePI{}}
\label{ssec:recurive_pi_nets}

\modelnamePI{} is a family of architectures that are high-degree polynomial expansions~\citep{chrysos2020pinets}. To reduce the parameters and enable the implementation of the polynomial expansion, coupled tensor decompositions are utilized. This results in a simple recursive formulation that enables an arbitrary degree of expansion. For instance, the \(N^{\text{th}}\) degree polynomial used for image recognition is expressed as: 
\begin{equation}
\boutvar_{1}=\left(\bm{H}_{[1]}^{T} \binvar\right) *\left(\bm{K}_{[1]}^{T} \bm{k}_{[1]}\right),
\end{equation}
\begin{equation}
\boutvar_{n}=\left(\bm{H}_{[n]}^{T} \binvar\right) *\left(\bm{J}_{[n]}^{T} \boutvar_{n-1}+\bm{K}_{[n]}^{T} \bm{k}_{[n]}\right)+\boutvar_{n-1},
\end{equation}
\begin{equation}
\boutvar=\bm{B} \boutvar_{N} + \bm{\theta},
\end{equation}
for $n=2, \ldots, N$. The symbol $\binvar$ is the input vector of the polynomial, $\boutvar$ is the output. The parameters $\bm{B}, \bm{\theta}, \{\bm{H}_{[n]}, \bm{J}_{[n]}, \bm{K}_{[n]}, \bm{k}_{[n]} \}_{n=1}^{N}$ are trainable. 
The aforementioned models can be used both in a hybrid setting (i.e., using polynomial expansion with element-wise activation) functions or as polynomial expansions. In the latter case, it was reported that despite the training accuracy reaching $100\%$, the testing accuracy was reduced when compared to \shortneuralnets.

\section{CCP-equivalent for the regularized model}
\label{ssec:CCP_r_pi_nets}
Beyond the aforementioned model of sec.~\ref{ssec:normalization_poly_proposed_model}, by changing the assumptions behind the tensor decomposition, one could retrieve another architecture. 
In this section, we demonstrate how an alternative parametrization, called CCP in \cite{chrysos2020pinets} can be reformulated in our context. 
A coupled CP decomposition (CCP) can be used for tensor parameters of PNs. The recursive equation of CCP can be expressed as:
$$
\boutvar_{1}=\bm{H}_{[1]}^{T} \binvar,
$$
$$
\boutvar_{n}=\left(\bm{H}_{[n]}^{T} \binvar\right) * \boutvar_{n-1}+\boutvar_{n-1},
$$
$$
\boutvar = \bm{B} \boutvar_{N}+\theta,
$$
for $n=1, \ldots, N$, the parameters $\bm{B} \in \mathbb{R}^{o \times r}, \bm{H}_{[n]} \in \mathbb{R}^{d \times r}$ for $n=1, \ldots, N$ are trainable. 

\begin{figure}[ht]
\vskip 0.2in
\begin{center}
\includegraphics[width=1.0\linewidth]{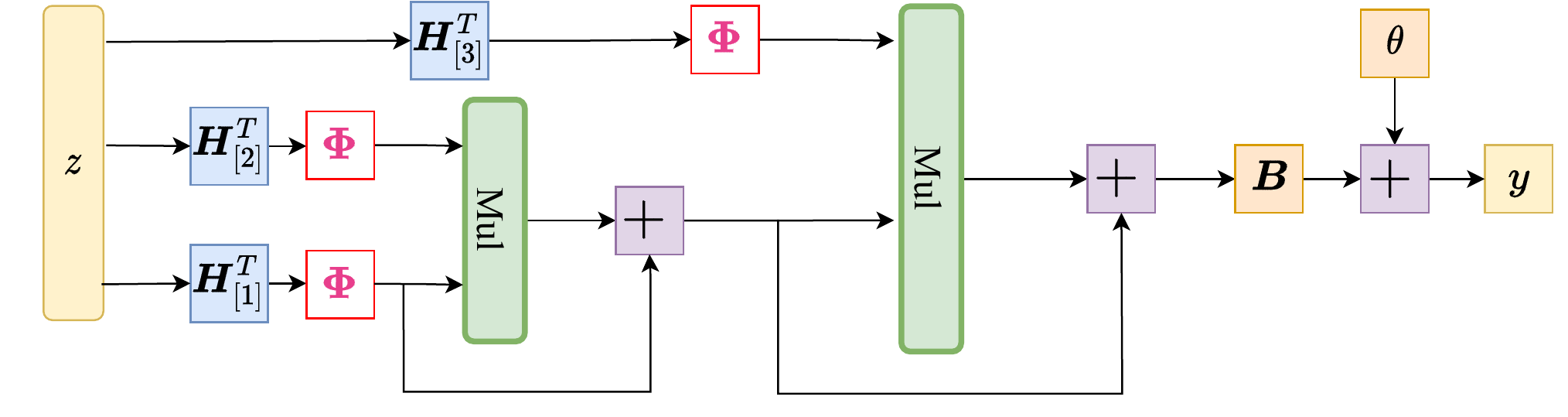}
\end{center}\vspace{-3mm}
\caption{Schematic illustration of the regularized CCP (for third degree approximation). Symbol $*$ refers to the Hadamard product. 
}
\label{fig:CCP}
\vskip -0.2in
\end{figure}

After introducing regularization matrix, $\bm{\Phi} \in \mathbb{R}^{r \times r}$, the modified recursive relationships for $n=2, \ldots, N$ can be expressed as follows:
$$
x_{n}=\left(\bm{\bm{\Phi}}\bm{H}_{[n]}^{T} \binvar\right) * \boutvar_{n-1}+\boutvar_{n-1}.
$$
A schematic
assuming a third order expansion ($N = 3$) is illustrated in Fig.~\ref{fig:CCP}.

\begin{table*}[ht]
\caption{Experimental settings in sec~\ref{ssec:normalization_poly_main_experiment_image_recognition} and sec~\ref{sec:imagenet}. Note the hyper-parameters of label smoothing are selected on the validation sets of \cifarten{} and \cifarhundred{}.
}
\label{tbl:experiment_setting}
\vskip 0.15in \vspace{-8mm}
\begin{center}
\small
\begin{tabular}{l|cc} 
& \cifarten{}/\cifarhundred{}/\stl{}/\tinyimagenet{} & ImageNet \\
\hline optimizer & SGD & SGD \\
base learning rate & $1 \mathrm{e}-1$ & $1 \mathrm{e}-1$ \\
weight decay & $5 \mathrm{e}-4$ & $1 \mathrm{e}-4$ \\
optimizer momentum & $0.9$ & $0.9$ \\
batch size & $128$ ($64$: \tinyimagenet{}) & $256$ \\
training epochs & $120$ & $100$ \\
learning rate schedule & multi-step decay & multi-step decay \\
 & exponential decay: \newmodelnamePI{}/ \modelnamedense{} (\tinyimagenet{}) &  \\
label smoothing: \newmodelnamePI{}/\modelnamedense{} & $0.1$: (\cifarten{} and \stl{})  & $0.1$ \\
 & $0.4$: (\cifarhundred{})  & \\
  & $0.6$: (\tinyimagenet{})  & \\
\end{tabular}
\end{center}
\end{table*}

\begin{table}[ht]
\caption{Experimental setting in sec~\ref{ssec:audio_classification}. 
}
\label{tbl:experiment_setting_speech}
\vskip 0.15in \vspace{-8mm}
\begin{center}
\small
\scalebox{0.92}{
\begin{tabular}{l|c} 
&  Speech Command \\
\hline optimizer & SGD \\
base learning rate & $1 \mathrm{e}-1$ \\
weight decay & $5 \mathrm{e}-4$ \\
optimizer momentum & $0.9$ \\
batch size & $128$ \\
training epochs & $120$ \\
learning rate schedule & multi-step decay \\
label smoothing: \newmodelnamePI{}/\modelnamedense{} & $0.1$  \\
\end{tabular}
}
\end{center}
\end{table}

\begin{table}[ht]
\caption{Experimental setting in sec~\ref{ssec:oxford_flower_classification}. 
}
\label{tbl:experiment_setting_oxford_flower}
\vskip 0.15in \vspace{-8mm}
\begin{center}
\small
\scalebox{0.92}{
\begin{tabular}{l|c} 
&  Oxford $102$ Flowers \\
\hline optimizer & SGD \\
base learning rate & $1 \mathrm{e}-1$ \\
weight decay & $5 \mathrm{e}-4$ \\
optimizer momentum & $0.9$ \\
batch size & $64$ \\
training epochs & $120$ \\
learning rate schedule & multi-step decay \\
label smoothing: \newmodelnamePI{}/\modelnamedense{} & $0.4$  \\
\end{tabular}
}
\end{center}
\end{table}

\section{Auxiliary tables and visualizations for experiments on the main paper}
\label{sec:normalization_poly_aux_tables_suppl}
Below we list the settings for experiments on the main paper in the Table~\ref{tbl:experiment_setting}, Table~\ref{tbl:experiment_setting_speech} and Table~\ref{tbl:experiment_setting_oxford_flower}. The Table~\ref{tbl:normalization_poly_ablation_results_varying_degrees_on_cifar10_cifar100_r_pi_net} ablates the accuracy for different degree polynomials on  \cifarten{} and \cifarhundred{}.

\begin{table}[t]
\caption{Accuracy of \newmodelnamePI{} with varying degree polynomials on \cifarten{} and \cifarhundred{}. Each block is a degree $2$ polynomial expansion, which results in the $2^6$ expansion if we add $6$ such blocks. Blocks with higher-degree can also be used, however we note that training those has not been as stable in our experience.} 
\label{tbl:normalization_poly_ablation_results_varying_degrees_on_cifar10_cifar100_r_pi_net}
\vskip 0.15in \vspace{-8mm}
\begin{center}
\begin{tabular}{|l|l|l|}
\hline
\bf Dataset           & \bf Degree polynomials & \bf Accuracy \\ \hline
\multirow{4}{*}{\cifarten{}} &    $2^2$ degree expansion                 &      $0.880 \pm 0.003$      \\ \cline{2-3} 
                  &          $2^4$ degree expansion           & $0.924 \pm 0.003$           \\ \cline{2-3} 
                  &          $2^6$ degree expansion           &  $0.931 \pm 0.001$          \\ \cline{2-3}
                  &          $2^8$ degree expansion           &   $0.945 \pm 0.000$    \\ \cline{2-3} 
                  &            $2^{10}$ degree expansion         &   $\bm{0.950 \pm 0.002}$        \\ \hline
\multirow{4}{*}{\cifarhundred{}} &    $2^2$ degree expansion                 &     $0.671 \pm 0.003$       \\ \cline{2-3} 
                  &          $2^4$ degree expansion           &    $0.732 \pm 0.002$        \\ \cline{2-3} 
                  &          $2^6$ degree expansion           & $0.738 \pm 0.002$          \\ \cline{2-3}
                  &          $2^8$ degree expansion         &    $0.769 \pm 0.002$            \\ \cline{2-3} 
                  &          $2^{10}$ degree expansion         &    $\bm{0.775 \pm 0.002}$        \\ \hline
\end{tabular}
\end{center}\vspace{-3mm}
\end{table}

\section{Additional experimental results}
\label{sec:normalization_poly_additional_experiments_suppl}

The following additional experimental results are added below:
\begin{enumerate}
    \item We evaluate the classification under limited training data in sec.~\ref{ssec:normalization_poly_suppl_limited_data}. 
    \item We conduct an error analysis for best and worst performing classes in sec.~\ref{ssec:normalization_poly_suppl_error_analysis}.
    \item In sec.~\ref{ssec:normalization_poly_suppl_conv_kernels}, an comparison with convolutional kernel networks is conducted. 
\end{enumerate}

Below, we also add details on the datasets used in this paper:

\textbf{Datasets}: The following datasets are used in our evaluation:
\begin{enumerate}
    \item \emph{\cifarten{}} \citep{krizhevsky2009learning} is a popular image recognition dataset consisting of $50,000$ training and $10,000$ testing images evenly distributed across $10$ classes. Each image is of resolution $32\times 32$.  
    \item \emph{\cifarhundred{}} \citep{cifar100} includes images similar to \cifarten. \cifarhundred{} contains $100$ object classes with $600$ ($500$ for training, $100$ for testing) images annotated per class. 
    \item \emph{\stl{}} \citep{coates2011analysis} contains $10$ object classes that are similar to \cifarten. Each image is of resolution $96\times 96$, while the dataset contains $5,000$ images. This dataset is used to evaluate the performance on images of higher resolution, while using limited data.
    \item \emph{\tinyimagenet{}} \citep{le2015tiny} contains $200$ object classes, where each image is of resolution $64\times 64$. There are $500$ images annotated per class, while the object classes demonstrate a larger variance than the aforementioned datasets. 
    \item \emph{Speech Commands dataset}~\citep{warden2018speech} includes $60, 000$ audio files; each audio contains a single word of a duration of one second. There are $35$ different words (classes) with each word having $1,500 - 4,100$ recordings. Every audio file is converted into a mel-spectrogram of resolution $32 \times 32$. 
    \item \emph{\imagenet{}}~\citep{deng2009imagenet} contains over one million training images and $50,000$ validation images from $1,000$ object classes. Each image depicts natural scenes and is annotated with a single object per image. 
\end{enumerate}

\subsection{Image classification with limited data}
\label{ssec:normalization_poly_suppl_limited_data}

\begin{figure}[tb]
    \centering
    \includegraphics[width=0.5\textwidth]{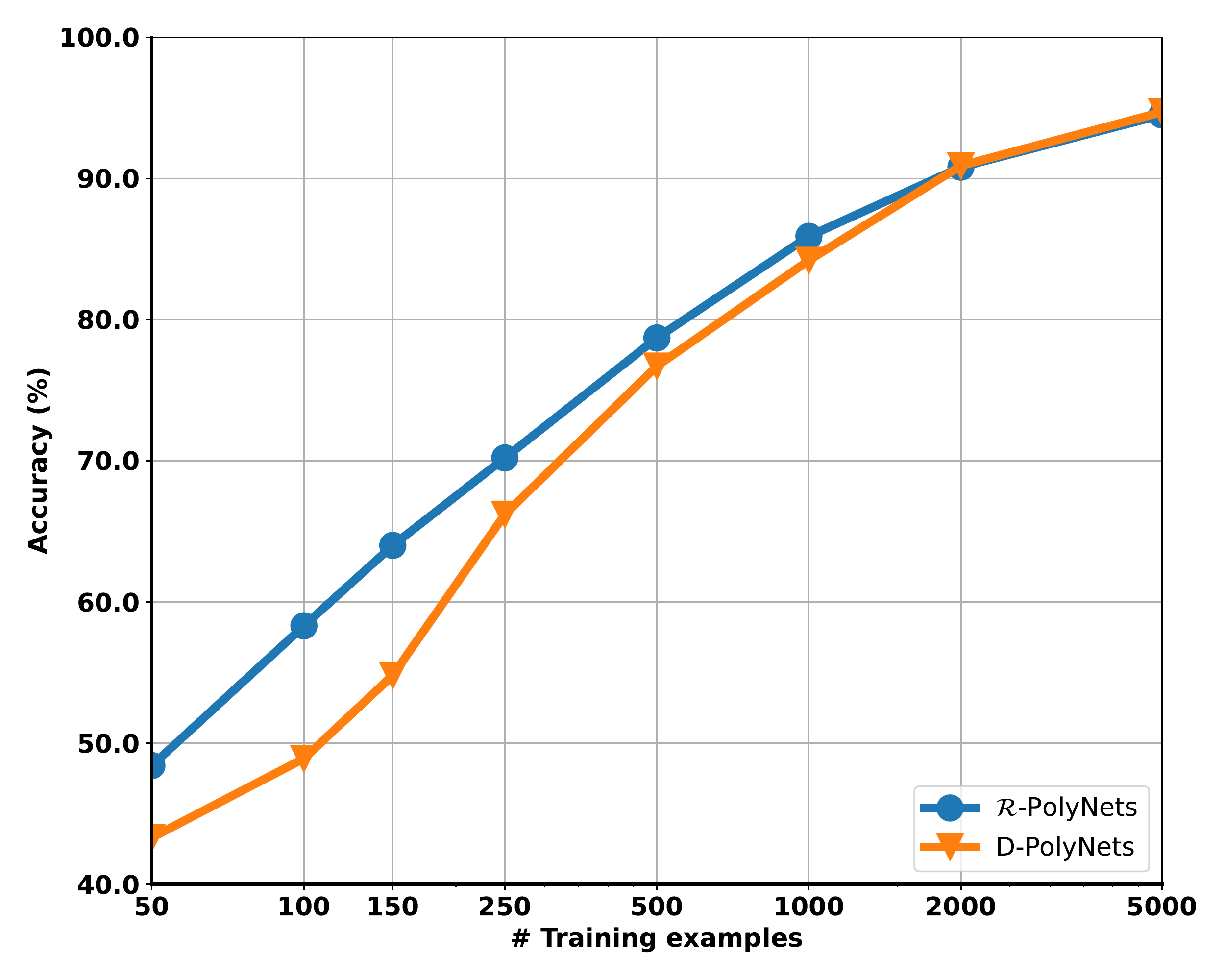}
    \caption{Image classification with limited data on \cifarten. The x-axis declares the number of training samples per class (log-axis). 
    }
    \label{fig:limit_number}
\end{figure}

We conduct an experiment on \cifarten{} in the presence of limited data. The hyper-parameters in sec.~\ref{ssec:normalization_poly_main_experiment_image_recognition} are used unchanged, while only the number of training samples of each class is reduced. The results in Table~\ref{tbl:results_limited_data} exhibit that \newmodelnamePI{} outperform \modelnamePI{} in the presence of limited training data. Notice that in the extreme case of only $50$ samples per class, there is a relative increase of $50\%$ from the accuracy of \modelnamePI. The goal of this experiment is to explore how \modelnamePI{} and \modelnamedense{} perform in the presence of limited data. Indeed, Fig.~\ref{fig:limit_number} confirms that both networks perform reasonably in the case of limited data. 

\begin{table}[ht]
\caption{Accuracy of image classification with limited data on \cifarten{}. Note that \newmodelnamePI{} without activation functions can outperform \modelnamePI{} without activation functions significantly on limited data of \cifarten{}. 
}
\label{tbl:results_limited_data}
\begin{center}
\vspace{-4mm}
\small
\scalebox{0.95}{
\begin{tabular}{|l|l|l|}
\hline
Training samples per class & \modelnamePI{} &\newmodelnamePI{}\\ \hline
  $50$      &    $0.314 \pm 0.005$  &  $\bm{0.484 \pm 0.004}$         \\ \hline
  $100$   &    $0.355 \pm 0.010$  &  $\bm{0.583 \pm 0.003}$\\ \hline
  $150$   &  $0.396 \pm 0.010$    &  $\bm{0.640 \pm 0.006}$ \\ \hline

\end{tabular}
}
\end{center}
\end{table}

\subsection{Error Analysis}
\label{ssec:normalization_poly_suppl_error_analysis}

\begin{figure}[htbp]
    \centering
    \subfloat[\small Most accurate class ($90.2\%$ accuracy)]{\includegraphics[width=0.5\linewidth]{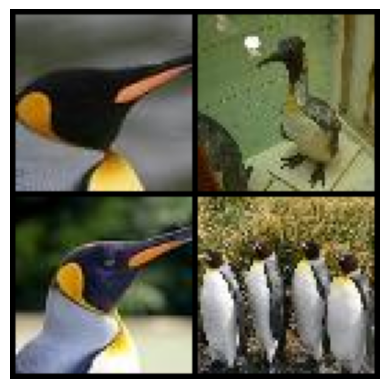}}
    \subfloat[\small Most misclassified class ($24.3\%$ accuracy)]{\includegraphics[width=0.5\linewidth]{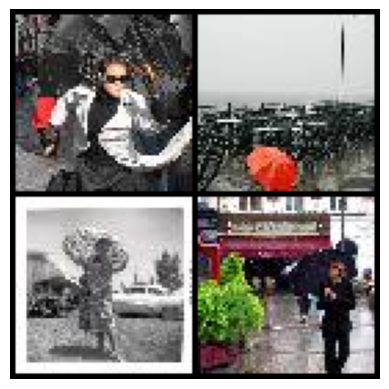}} \vspace{-0.1mm}
    \caption{Images of the Most Accurate/Misclassified Class recognized by \newmodelnamePI{}. As in Fig~\ref{fig:images_most_accurate_misclassified}, the king penguins in (a) have similar characteristics, and occupy most regions in the images. On the other side, the images in (b) are dominated by other objects such as persons and landscape. 
    }
    \label{fig:images_most_accurate_misclassified}
\end{figure}

We use our best-performing \newmodelnamePI{} to calculate per-class error rates for all 200 classes on the validation dataset of a large-scale classification dataset,  \tinyimagenet{} \citep{le2015tiny}. We report the top-5 accurate and misclassified classes in Table~\ref{tbl:top_5_accuract_inaccurate}. Also, we present the images of the most accurate class (king penguin) and the most misclassified class (umbrella) in Fig.~\ref{fig:images_most_accurate_misclassified}.

\begin{figure}[htbp]
    \centering
    \subfloat[\small Lesser panda (\modelnamePI{}: $92.7\%$ accuracy)
    ]{\includegraphics[width=0.5\linewidth]{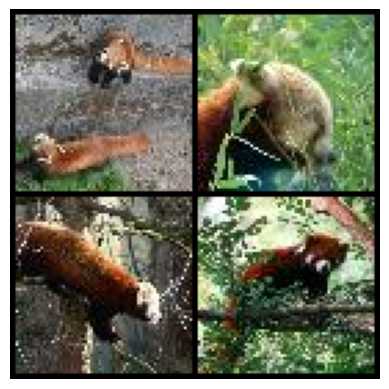}}
    \subfloat[\small saliency maps of \modelnamePI{}]{\includegraphics[width=0.5\linewidth]{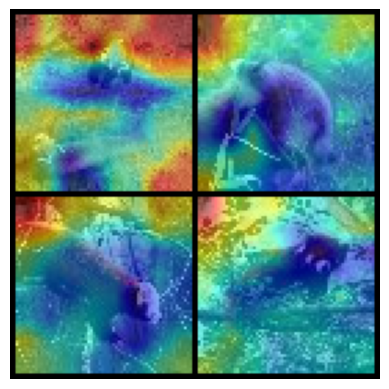}}  \vspace{-0.1mm}
    \subfloat[\small Lesser panda (\newmodelnamePI{}: $90.0\%$ accuracy)]{\includegraphics[width=0.5\linewidth]{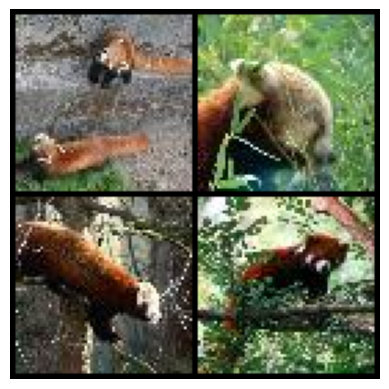}}
    \subfloat[\small saliency maps of \newmodelnamePI{}]{\includegraphics[width=0.5\linewidth]{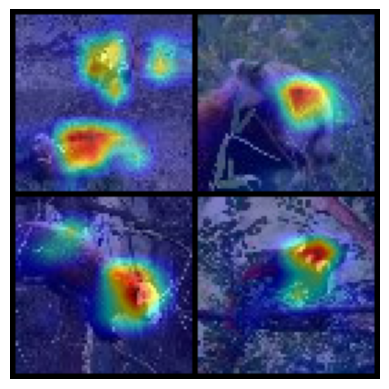}}
    \caption{Saliency maps of \modelnamePI{} and \newmodelnamePI{}. As in Fig~\ref{fig:saliency_maps}, \modelnamePI{} can recognize the lesser panda in (a) at $93\%$ accuracy, but the saliency maps in (b) indicate \modelnamePI{} can not concentrate on the main object in an image. The saliency maps in (d) indicate \newmodelnamePI{} can concentrate on the main object in an image.}
    \label{fig:saliency_maps}
\end{figure}

Remarkably, \newmodelnamePI{} achieve above $85\%$ validation accuracy for the top-5 accurate classes. We analyse the images of the most accurate class and misclassified class. As shown in Fig.~\ref{fig:images_most_accurate_misclassified}, the king penguins occupy most regions in the images. Also, they have similar shape, color and texture. On the other side, the umbrellas in Fig.~\ref{fig:images_most_accurate_misclassified} have different colors and shapes. Furthermore, the images are dominated by the other objects such as human beings and landscape. The saliency maps in Fig.~\ref{fig:saliency_maps} computed by GradCAM~\citep{selvaraju2017grad} indicate \newmodelnamePI{} can concentrate on the main object in an image. By comparison, \modelnamePI{} recognize the lesser panda at $92.7\%$ accuracy, but the saliency maps in Fig.~\ref{fig:saliency_maps} show \modelnamePI{} do not concentrate on the main object in an image.

\begin{table}[ht]
\caption{Top-5 Accurate/Misclassified Classes on \tinyimagenet{}. Note that \newmodelnamePI{} can achieve above $85\%$ validation accuracy for the top-5 accurate classes.}
\label{tbl:top_5_accuract_inaccurate}
\vskip 0.15in \vspace{-8mm}
\begin{center}
\small
\begin{tabular}{l|l|l|l}
\hline Class Name & Accuracy & Class Name & Accuracy \\
\hline king penguin & $0.902$ & backpack & $0.358$ \\
\hline lesser panda & $0.900$ & bucket & $0.327$ \\
\hline sea slug & $0.895$ & plunger & $0.296$ \\
\hline bullet train & $0.882$ & wooden spoon & $0.278$ \\
\hline Persian cat & $0.879$ & umbrella & $0.243$ \\
\hline
\end{tabular}
\end{center} 
\end{table}

\subsection{Comparison with convolutional kernel networks}
\label{ssec:normalization_poly_suppl_conv_kernels}
We conduct experiments to compare \newmodelnamePI{} and \modelnamedense{} with supervised convolutional kernel networks (SCKNs)~\citep{mairal2016end}, which is among the principled design choices. The results of \newmodelnamePI{} and \modelnamedense{} are the same as those in the main paper, i.e., in Table~\ref{tbl:results}. The accuracy for each model is reported in Table~\ref{tbl:results_ckn}. Notice that the proposed \newmodelnamePI{} and \modelnamedense{} outpuperform the newly added baseline.

\begin{table}[t]
\caption{Accuracy on \cifarten{}, \cifarhundred{}, \stl{} and \tinyimagenet{}. The symbol ‘\# par’ abbreviates the number of parameters. \modelnamedense{} containing 7M parameters. Note that \newmodelnamePI{} and \modelnamedense{} without activation functions can outperform SCKNs on \cifarten{}, \cifarhundred{}, \stl{} and \tinyimagenet{} by a large margin. 
} 
\label{tbl:results_ckn}
\vskip 0.15in \vspace{-6mm}
\begin{center}
\small
\begin{tabular}{llll}
\multicolumn{1}{c}{\bf Dataset}  &\multicolumn{1}{c}{\bf Model} &\multicolumn{1}{c}{\bf \# par} &\multicolumn{1}{c}{\bf Accuracy}
\\ \hline
\multirow{3}{*}{\cifarten{}} 
& SCKNs & $3.4$M  &  $0.895 \pm 0.002$       \\
& \newmodelnamePI{} & $11.9$M  & $0.945 \pm 0.000$ \\
& \modelnamedense{} & $7.1$M & $\bm{0.947 \pm 0.002}$        \\
\cline{1-4}
\multirow{3}{*}{\cifarhundred{}} 
& SCKNs &  $3.5$M  & $0.610 \pm 0.003$        \\
& \newmodelnamePI{} & $11.9$M  & $\bm{0.769 \pm 0.002}$ \\
& \modelnamedense{} & $7.2$M &  $0.767 \pm 0.003$        \\  

\cline{1-4}
\multirow{3}{*}{\stl{}} 
& SCKNs & $3.4$M  &  ${0.527\pm0.012}$        \\
& \newmodelnamePI{} & $11.9$M  & $0.828 \pm 0.003$ \\
& \modelnamedense{} & $7.1$M &  $\bm{0.834 \pm 0.006}$        \\  

\cline{1-4}
\multirow{3}{*}{\tinyimagenet{}} 
& SCKNs & $4.2$M  & $0.409 \pm 0.001$         \\
& \newmodelnamePI{} & $12.0$M  & $0.615 \pm 0.004$ \\
& \modelnamedense{} & $7.2$M &  $\bm{0.618 \pm 0.001}$        \\  

\cline{1-4}

\end{tabular}
\end{center}
\vskip -0.1in
\end{table}

\subsection{Regularized PDC, ResNext and dense connections for PDC}
To showcase the representative power of the regularized polynomial expansion and dense connections across different polynomial nets, we firstly apply the proposed regularization schemes (IBN + max pooling + Dropblock + Label smoothing) in the influential ResNext~\cite{xie2017aggregated} and the recent PDC~\cite{chrysos2022polynomial}. The regularized ResNext is called $\mathcal{R}$-PolyNeXt, while the regularized PDC is called $\mathcal{R}$-PDC. The results for ResNext are reported in Table~\ref{tbl:norm_poly_resnext_results}. Even though $\mathcal{R}$-PolyNeXt performs on par with ResNext, we notice that there is some training instability that did not emerge in regularizing PDC or \modelnamePI. It is possible that further tuning is required for converting more complex models, such as ResNext, into polynomial expansions. 

\begin{figure}[]
\vskip -0.22in
\begin{center}
\includegraphics[width=\linewidth]{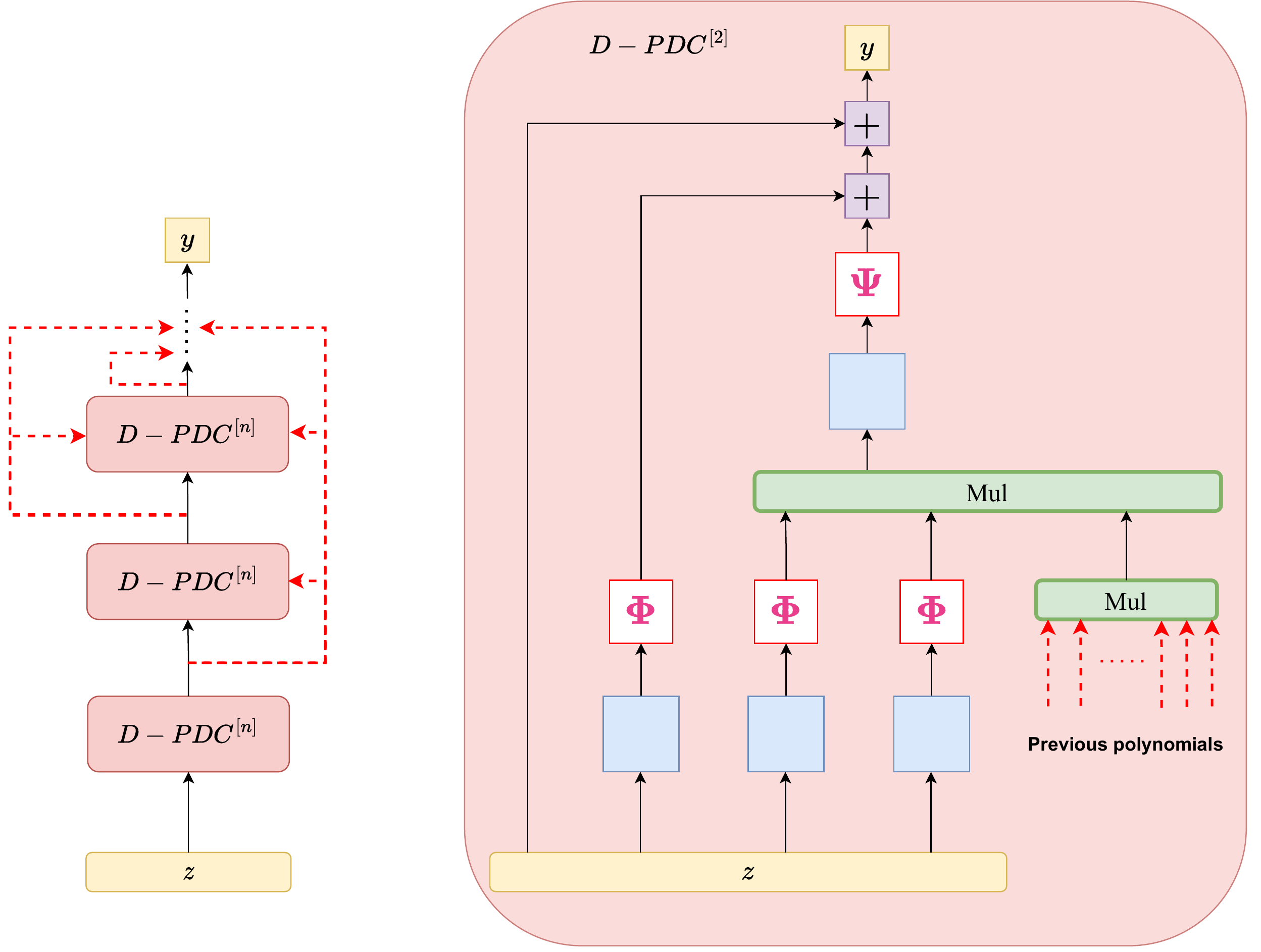}
\end{center}\vspace{-3mm}
\caption{Schematic illustration of $\mathcal{D}$-PDC. On the left the overall structure is presented, while on the right a single second-degree polynomial using the structure of $\mathcal{D}$-PDC is visualized. The red arrows depict the newly added connections with respect to previous polynomial expansions. 
}
\label{fig:schematic_D_PDC}
\end{figure}

Furthermore, we also enable additional skip connections across polynomials for PDC. The new type of PDC is called $\mathcal{D}$-PDC. The schematic in Fig.~\ref{fig:schematic_D_PDC} depicts $\mathcal{D}$-PDC assuming each polynomial includes a single recursive step. This can be trivially extended to any number of recursive steps, while each polynomial can also rely on a different tensor decomposition. The same regularization scheme (IBN + max pooling + Dropblock + Label smoothing) in \modelnamedense{} is used in $\mathcal{D}$-PDC. The accuracy for each model is reported in Table~\ref{tbl:results_DPDC}. Notice that the $\mathcal{R}$-PDC and $\mathcal{D}$-PDC both outperform the PDC. 
The rest of the patterns, e.g., $\mathcal{D}$-PDC versus $\mathcal{R}$-PDC, are similar to the experiments in the main paper.

\begin{table}[t]
\caption{Accuracy on \cifarten{}, \cifarhundred{}, \stl{} and \tinyimagenet{}. The symbol ‘\# par’ abbreviates the number of parameters. Note that $\mathcal{R}$-PDC and $\mathcal{D}$-PDC without activation functions can outperform PDC without activation functions significantly on \cifarten{}, \cifarhundred{}, \stl{} and \tinyimagenet{}.
} 
\label{tbl:results_DPDC}
\vskip 0.15in \vspace{-6mm}
\begin{center}
\small
\begin{tabular}{llll}
\multicolumn{1}{c}{\bf Dataset}  &\multicolumn{1}{c}{\bf Model} &\multicolumn{1}{c}{\bf \# par} &\multicolumn{1}{c}{\bf Accuracy}
\\ \hline
\multirow{3}{*}{\cifarten{}} 
& {PDC} & {$5.4$M} & {$0.909 \pm 0.002$}        \\
& $\mathcal{R}$-PDC & $7.3$M  &  $0.947 \pm 0.001$      \\
& $\mathcal{D}$-PDC & $6.0$M  & $\bm{0.949 \pm 0.002}$        \\
\cline{1-4}
\multirow{3}{*}{\cifarhundred{}} 
& {PDC} & {$5.5$M} & {$0.689 \pm 0.002$}        \\
& $\mathcal{R}$-PDC & $7.4$M  & $0.757 \pm 0.003$        \\
& $\mathcal{D}$-PDC & $6.0$M  & $\bm{0.762 \pm 0.001}$        \\

\cline{1-4}
\multirow{3}{*}{\stl{}} 
& {PDC} & {$5.4$M} &   {${0.681\pm0.006}$}     \\
& $\mathcal{R}$-PDC & $7.3$M  & ${0.833\pm0.007}$       \\
& $\mathcal{D}$-PDC & $6.0$M  & $\bm{0.855 \pm 0.003}$        \\

\cline{1-4}
\multirow{3}{*}{\tinyimagenet{}} 
& {PDC} & {$5.5$M} & {$0.452 \pm 0.002$}       \\
& $\mathcal{R}$-PDC & $7.4$M  & $0.560 \pm 0.005$        \\
& $\mathcal{D}$-PDC & $6.0$M  &  $\bm{0.569 \pm 0.002}$       \\

\cline{1-4}

\end{tabular}
\end{center}
\vskip -0.1in
\end{table}

\begin{table}[t]
\caption{Accuracy of \modelnamedense{} without IBN and without label smoothing (mentioned as `\modelnamedense{} without reg' below) on \cifarten{} and \cifarhundred{}. The symbol ‘\# par’ abbreviates the number of parameters.
} 
\label{tbl:dense_pi_net_without_reg_results}
\vskip 0.15in \vspace{-6mm}
\begin{center}
\small
\begin{tabular}{llll}
\multicolumn{1}{c}{\bf Dataset}  &\multicolumn{1}{c}{\bf Model} &\multicolumn{1}{c}{\bf \# par} &\multicolumn{1}{c}{\bf Accuracy}
\\ \hline
\multirow{2}{*}{\cifarten{}} &
\multicolumn{1}{l}{{\modelnamePI{}}}   & {$11.9$M} & \multicolumn{1}{l}{{$0.907 \pm 0.003$}} \\\cline{2-4}
& \modelnamedense{} without reg & $7.1$M & $\bm{0.934 \pm 0.002}$        \\
\cline{1-4}
\multirow{2}{*}{\cifarhundred{}} &
\multicolumn{1}{l}{{\modelnamePI{}}}               & {$11.9$M}            & \multicolumn{1}{l}{{$0.677 \pm 0.006$}} \\\cline{2-4} 
& \modelnamedense{} without reg  & $7.2$M &  $\bm{0.726 \pm 0.006}$        \\  
\cline{1-4}

\end{tabular}
\end{center}
\vskip -0.1in
\end{table}

\begin{table}[t]
\caption{Accuracy of ResNext~\cite{xie2017aggregated} and the corresponding $\mathcal{R}$-PolyNeXt on \cifarten{} and \cifarhundred{}.} 
\label{tbl:norm_poly_resnext_results}
\vskip 0.15in \vspace{-6mm}
\begin{center}
\small
\begin{tabular}{llll}
\multicolumn{1}{c}{\bf Dataset}  &\multicolumn{1}{c}{\bf Model} &\multicolumn{1}{c}{\bf \# par} &\multicolumn{1}{c}{\bf Accuracy}
\\ \hline
\multirow{2}{*}{\cifarten{}} &
\multicolumn{1}{l}{\textcolor{baseline_color}{ResNeXt-$29$, $8 \times 64$d}}   & \textcolor{baseline_color}{$34.4$M} & \multicolumn{1}{l}{\textcolor{baseline_color}{$0.964$}} \\\cline{2-4}
&  $\mathcal{R}$-PolyNeXt-$29$, $8 \times 64$d & $38.6$M & $\bm{0.965}$        \\
\cline{1-4}
\multirow{2}{*}{\cifarhundred{}} &
\multicolumn{1}{l}{\textcolor{baseline_color}{ResNeXt-$29$, $8 \times 64$d}}               & \textcolor{baseline_color}{$34.4$M}            & \multicolumn{1}{l}{\textcolor{baseline_color}{$0.822$}} \\\cline{2-4} 
& $\mathcal{R}$-PolyNeXt-$29$, $8 \times 64$d  & $38.7$M &  $\bm{0.824}$        \\  
\cline{1-4}

\end{tabular}
\end{center}
\vskip -0.1in
\end{table}

\subsection{Comparison with deeper ResNets}

We conduct experiments to compare deeper \newmodelnamePI{} and \modelnamedense{} with deeper ResNets. The experimental settings described in sec.~\ref{ssec:normalization_poly_main_experiment_image_recognition} remain unchanged for these comparisons. The accuracy for each model is reported in Table~\ref{tbl:results_larger}. It is noteworthy that the proposed \newmodelnamePI{} and \modelnamedense{} outperform ResNets when their architectures are deeper.

\begin{table}[t]
\caption{Accuracy on \cifarten{} and \cifarhundred{}. The symbol ‘\# par’ abbreviates the number of parameters. Note that deeper \newmodelnamePI{} and \modelnamedense{} without activation functions can outperform deeper ResNets on \cifarten{}, and \cifarhundred{}. 
} 
\label{tbl:results_larger}
\vskip 0.15in \vspace{-6mm}
\begin{center}
\small
\begin{tabular}{llll}
\multicolumn{1}{c}{\bf Dataset}  &\multicolumn{1}{c}{\bf Model} &\multicolumn{1}{c}{\bf \# par} &\multicolumn{1}{c}{\bf Accuracy}
\\ \hline
\multirow{3}{*}{\cifarten{}} 
& \multicolumn{1}{l}{\textcolor{baseline_color}{ResNet34}}   & \textcolor{baseline_color}{$21.3$M} & \multicolumn{1}{l}{\textcolor{baseline_color}{$0.947 \pm 0.002$}} \\
& \newmodelnamePI{}34 & $22.5$M  & $0.950 \pm 0.001$ \\
& \modelnamedense{}34 & $13.5$M & $\bm{0.951 \pm 0.002}$        \\
\cline{2-4}
& \multicolumn{1}{l}{\textcolor{baseline_color}{ResNet152}}   & \textcolor{baseline_color}{$58.2$M} & \multicolumn{1}{l}{\textcolor{baseline_color}{$0.943 \pm 0.003$}} \\
& \newmodelnamePI{}152 & $58.5$M  & $0.952 \pm 0.001$ \\
& \modelnamedense{}152 & $54.2$M & $\bm{0.953 \pm 0.002}$        \\
\cline{1-4}
\multirow{3}{*}{\cifarhundred{}} 
& \multicolumn{1}{l}{\textcolor{baseline_color}{ResNet34}}   & \textcolor{baseline_color}{$21.3$M} & \multicolumn{1}{l}{\textcolor{baseline_color}{$0.762 \pm 0.004$}} \\
& \newmodelnamePI{}34 & $22.6$M  & $\bm{0.788 \pm 0.002}$ \\
& \modelnamedense{}34 & $13.5$M & $0.787 \pm 0.001$        \\
\cline{2-4}
& \multicolumn{1}{l}{\textcolor{baseline_color}{ResNet152}}   & \textcolor{baseline_color}{$58.3$M} & \multicolumn{1}{l}{\textcolor{baseline_color}{$0.768 \pm 0.005$}} \\
& \newmodelnamePI{}152 & $58.5$M  & $\bm{0.793 \pm 0.004}$ \\
& \modelnamedense{}152 & $54.2$M & $0.791 \pm 0.001$        \\
\cline{1-4}

\end{tabular}
\end{center}
\vskip -0.1in
\end{table}

\subsection{FLOPs}

We compute the floating-point operations per second (FLOPs) for \newmodelnamePI{}, \modelnamedense{}, \modelnamePI{}, and ResNet18 on both small datasets and ImageNet. The results of these computations are presented in Table~\ref{tbl:small_datasets_flops} and Table~\ref{tbl:imagenet_flops}. Notice that the proposed \newmodelnamePI{} has a similar FLOP as the previously proposed \modelnamePI, while \modelnamedense{} has only a marginal increase in the FLOPs.

\begin{table}[t]
\caption{FLOPs on \cifarten{}, \cifarhundred{}, \stl{} and \tinyimagenet{}. 
} 
\label{tbl:small_datasets_flops}
\vskip 0.15in \vspace{-6mm}
\begin{center}
\small
\begin{tabular}{lll}
\multicolumn{1}{c}{\bf Dataset}  &\multicolumn{1}{c}{\bf Model} &\multicolumn{1}{c}{\bf GFLOPs}
\\ \hline
\multirow{2}{*}{\cifarten{}} &
\multicolumn{1}{l}{\textcolor{baseline_color}{ResNet18}}   & \textcolor{baseline_color}{$0.56$} \\
&   \multicolumn{1}{l}{\textcolor{baseline_color}{Hybrid \modelnamePI{}}} & \textcolor{baseline_color}{$0.46$}  \\\cline{2-3}
& \multicolumn{1}{l}{\modelnamePI{}} & $0.59$  \\ 
& \newmodelnamePI{} & $0.59$  \\
& \modelnamedense{} & $0.55$ \\
\cline{1-3}
\multirow{3}{*}{\cifarhundred{}} &
\multicolumn{1}{l}{\textcolor{baseline_color}{ResNet18}}               & \textcolor{baseline_color}{$0.56$}            \\
& \multicolumn{1}{l}{\textcolor{baseline_color}{Hybrid \modelnamePI{}}}     &     \textcolor{baseline_color}{$0.46$}        \\\cline{2-3} 
& \multicolumn{1}{l}{\modelnamePI{}} & $0.59$\\ 
& \newmodelnamePI{} & $0.59$  \\
& \modelnamedense{} & $0.55$ \\  

\cline{1-3}
\multirow{3}{*}{\stl{}} &
\multicolumn{1}{l}{\textcolor{baseline_color}{ResNet18}}                &   \textcolor{baseline_color}{$5.01$}   \\
& \multicolumn{1}{l}{\textcolor{baseline_color}{Hybrid \modelnamePI{}}}        &   \textcolor{baseline_color}{$4.11$}           \\\cline{2-3} 
& \multicolumn{1}{l}{\modelnamePI{}} & $5.31$ \\ 
& \newmodelnamePI{} & $5.31$ \\
& \modelnamedense{} & $4.94$ \\  

\cline{1-3}
\multirow{3}{*}{\tinyimagenet{}} &
\multicolumn{1}{l}{\textcolor{baseline_color}{ResNet18}}                  &   \textcolor{baseline_color}{$2.23$}    \\ 
&  \multicolumn{1}{l}{\textcolor{baseline_color}{Hybrid \modelnamePI{}}}    & \textcolor{baseline_color}{$1.83$} \\ \cline{2-3} 
& \multicolumn{1}{l}{\modelnamePI{}} & $2.36$  \\ 
& \newmodelnamePI{} & $2.36$  \\
& \modelnamedense{} & $2.19$ \\  

\cline{1-3}

\end{tabular}
\end{center}
\vskip -0.1in
\end{table}

\begin{table}[t]
\caption{FLOPs on ImageNet. Notice that the proposed \newmodelnamePI{} has a similar FLOP as the previously proposed \modelnamePI.}
\label{tbl:imagenet_flops}
\vskip 0.15in \vspace{-8mm}
\begin{center}
\small
\begin{tabular}{ll}
\hline Model & GFLOPs \\
\hline ImageNet-1K trained models &  \\
\hline ResNet18 & $1.82$ \\
ResNet18 without activations & $1.82$ \\
Hybrid \modelnamePI{} &  $1.92$ \\
\modelnamePI{} & $1.92$ \\
\rowcolor{Gray}
\newmodelnamePI{} & $1.92$ \\
\rowcolor{Gray}
\modelnamedense{} & $1.98$ \\
\hline
\end{tabular}
\end{center}
\end{table}

\end{document}